\documentclass[lettersize,journal]{IEEEtran}
\usepackage{amsmath,amsfonts}
\usepackage{algorithmic}
\usepackage{algorithm}
\usepackage{array}
\usepackage{textcomp}
\usepackage{stfloats}
\usepackage{url}
\usepackage{verbatim}
\usepackage{graphicx}
\usepackage{cite}
\usepackage{tabularx, xcolor, colortbl, booktabs, multirow, array} 
\newcolumntype{C}{>{\centering\arraybackslash}X} 
\usepackage[misc]{ifsym}
\usepackage[table]{xcolor}
\usepackage{makecell}
\usepackage{threeparttable}
\usepackage{pifont}
\usepackage{subcaption}  
\usepackage{hyperref}

\usepackage{tcolorbox}
\tcbuselibrary{listings, breakable}

\usepackage{listings}
\newtcolorbox{promptbox}{
  breakable,
  colback=gray!3,
  colframe=gray!40,
  boxrule=0.4pt,
  arc=2pt,
  left=6pt,
  right=6pt,
  top=6pt,
  bottom=6pt
}

\begin{document}
\title{FaithSCAN: Model-Driven Single-Pass Hallucination Detection for Faithful Visual Question Answering}
\author{Chaodong Tong, Qi Zhang, Chen Li, Lei Jiang, and Yanbing Liu*
\IEEEcompsocitemizethanks{
\IEEEcompsocthanksitem Chaodong Tong, Yanbing Liu are with the Institute of Information Engineering, Chinese Academy of Sciences (CAS) and the School of Cyber Security, University of CAS, Beijing 100093, China. E-mail: \{tongchaodong, liuyanbing\}@iie.ac.cn.
\IEEEcompsocthanksitem Qi Zhang is with the China Industrial Control Systems Cyber Emergency Response Team, Beijing 100040, China. E-mail: bonniezhangqi@126.com.
\IEEEcompsocthanksitem Chen Li is with the China Electronics Standardization Institute, Ministry of Industry and Information Technology of the People’s Republic of China, Beijing 100007, China. E-mail: lichen@cesi.cn.
\IEEEcompsocthanksitem Lei Jiang is with the Institute of Information Engineering, Chinese Academy of Sciences (CAS), Beijing 100093, China. E-mail: jianglei@iie.ac.cn.
\IEEEcompsocthanksitem *Corresponding author: Yanbing Liu. 
}
}


\maketitle

\begin{abstract}
Faithfulness hallucinations in VQA occur when vision-language models produce fluent yet visually ungrounded answers, severely undermining their reliability in safety-critical applications. Existing detection methods mainly fall into two categories: external verification approaches relying on auxiliary models or knowledge bases, and uncertainty-driven approaches using repeated sampling or uncertainty estimates. The former suffer from high computational overhead and are limited by external resource quality, while the latter capture only limited facets of model uncertainty and fail to sufficiently explore the rich internal signals associated with the diverse failure modes. Both paradigms thus have inherent limitations in efficiency, robustness, and detection performance.
To address these challenges, we propose FaithSCAN: a lightweight network that detects hallucinations by exploiting rich internal signals of VLMs, including token-level decoding uncertainty, intermediate visual representations, and cross-modal alignment features. These signals are fused via branch-wise evidence encoding and uncertainty-aware attention. We also extend the LLM-as-a-Judge paradigm to VQA hallucination and propose a low-cost strategy to automatically generate model-dependent supervision signals, enabling supervised training without costly human labels while maintaining high detection accuracy.
Experiments on multiple VQA benchmarks show that FaithSCAN significantly outperforms existing methods in both effectiveness and efficiency. In-depth analysis shows hallucinations arise from systematic internal state variations in visual perception, cross-modal reasoning, and language decoding. Different internal signals provide complementary diagnostic cues, and hallucination patterns vary across VLM architectures, offering new insights into the underlying causes of multimodal hallucinations.
\end{abstract}


\section{Introduction}
\begin{figure*}[!htbp]
    \centering
    \includegraphics[width=\textwidth]{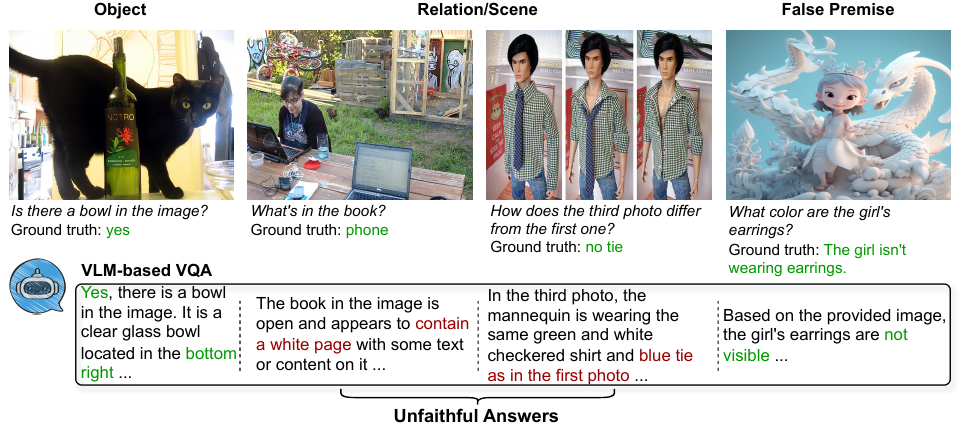}
    
    \caption{Illustration of three representative VQA scenarios: (a) object existence questions, (b) relationships between two objects within a single image, and (c) reasoning with a false-premise question. Modern VLM-based QA systems perform reliably on simple questions, but often produce unfaithful or hallucinated answers in complex scenarios.}

    \label{fig:fig1}
\end{figure*}

Visual Question Answering (VQA) is a representative image-to-text (I2T) task in which a model generates natural language answers given an image and a corresponding question~\cite{huang2025survey}. Recent advances in multimodal large language models (MLLMs) and vision-language models (VLMs) have enabled impressive integration of visual perception and language understanding, supporting increasingly complex open-ended VQA tasks~\cite{ging2024openended,Yu2024TowardsOV}. Despite impressive performance, they frequently generate fluent yet ungrounded or incorrect responses, a phenomenon known as hallucination~\cite{ji2023survey,farquhar2024detecting,bai2024hallucination}, which in VQA settings manifests as responses misaligned with the input visual evidence or contradictory to factual knowledge~\cite{liu2024survey,hu2024mitigating}. Among them, \emph{faithfulness hallucinations} \cite{huang2025survey}, where answers violate visual grounding, are particularly critical in VQA, since reliable factual verification depends on the model first attending to and correctly interpreting the image~\cite{li-etal-2023-evaluating}. Consequently, detecting faithfulness hallucinations is a necessary step for ensuring answer reliability and user trust, motivating the development of effective detection methods that can identify unfaithful answers (examples illustrated in Fig.~\ref{fig:fig1}).

Motivated by the importance of faithfulness hallucination detection in VQA, a variety of detection methods have been proposed in recent years.
Existing approaches can be broadly grouped into two paradigms: \emph{external verification-based} methods and \emph{uncertainty-driven} methods~\cite{bai2024hallucination,chen2025survey}.
External verification approaches detect hallucinations by explicitly checking cross-modal consistency between the generated answer and the visual input, often relying on auxiliary components such as object detectors, VQA models, scene graphs, or large language model (LLM) judges~\cite{li-etal-2023-evaluating,jing2024faithscore,gu2024survey}.
By decomposing answers into atomic claims and validating them against visual evidence, these methods are capable of identifying fine-grained inconsistencies.
However, their reliance on additional pretrained models and multi-stage pipelines leads to high computational overhead and makes the final performance sensitive to error propagation from external components.
Moreover, such methods operate largely independently of the internal reasoning process of the target VLM, limiting their scalability and integration in real-world deployment~\cite{jing2024faithscore}.

In contrast, uncertainty-driven methods approach hallucination detection from the perspective of model confidence.
Empirical studies in both language models and vision--language models have shown that hallucinated outputs are often associated with elevated predictive uncertainty~\cite{huang2025survey,liu2024survey}.
Representative approaches, such as semantic entropy (SE)~\cite{farquhar2024detecting} and likelihood-based measures~\cite{farquhar2024detecting, zhang2024vl, ding2024hallu, tong2025semantic}, estimate hallucination likelihood through sampling model outputs under different decoding strategies, prompt reformulations, or visual perturbations, followed by entropy or distributional analysis. Recent work enhances uncertainty estimation by incorporating visual cues, such as vision-amplified semantic entropy~\cite{liao2025vision} and dense geometric entropy~\cite{gautam2025hedge}.
In addition, methods leveraging internal signals like attention patterns and intermediate embeddings are emerging. Approaches such as DHCP~\cite{zhang2025dhcp} and OPERA~\cite{huang2024opera} focus on attention dynamics to detect over-reliance on unverified features, while HallE-Control~\cite{zhai2023halle} modifies embeddings to suppress unreliable visual details. These methods highlight the value of integrating internal uncertainty signals.

Despite their effectiveness, existing uncertainty-based methods suffer from two key limitations.
\textbf{First, they rely on repeated sampling and generation-level statistics, resulting in high inference cost and unstable behavior.}
This dependence on stochastic decoding and prompt configurations significantly limits their practicality in latency-sensitive or large-scale VQA settings.
\textbf{Second, they fail to jointly leverage the heterogeneous uncertainty signals inherently encoded in modern VLMs.}
Uncertainty arising from language decoding, visual perception, and vision--language interaction is often treated independently or partially ignored, rather than being modeled in a unified manner.
As a result, the complementary information carried by different internal uncertainty sources remains largely unused for hallucination detection.

To tackle these limitations, we propose \textbf{FaithSCAN}, a lightweight faithfulness hallucination detection network for VQA. FaithSCAN treats hallucination detection as a supervised learning problem, where multiple sources of internal uncertainty are extracted from a single forward pass of a VLM. Specifically, it leverages three complementary uncertainty signals: (i) token-level predictive uncertainty from language decoding, (ii) intermediate visual representations reflecting the model’s perception of the image, and (iii) cross-modal aligned representations that project visual information into the textual semantic space. These heterogeneous signals are processed via \emph{Branch-wise Evidence Encoding} and aggregated using \emph{Uncertainty-Aware Attention}, enabling the detector to capture hallucination patterns across multiple reasoning pathways. Supervision is provided through a model-driven Visual-NLI procedure, extending the LLM-as-a-Judge paradigm to VQA, and is further validated by small-scale human verification, ensuring label reliability without requiring large-scale manual annotation.

Our contributions are summarized as follows:
\begin{itemize}
\item We introduce FaithSCAN, a novel hallucination detection architecture that leverages internal uncertainty signals from a single VLM forward pass.
\item FaithSCAN jointly models uncertainty from language decoding, visual perception, and cross-modal interaction through branch-wise encoding and uncertainty-aware aggregation, linking internal signals to hallucinations.
\item We propose a model-aware supervision strategy by extending the LLM-as-a-Judge paradigm to multimodal hallucination detection in VQA, where hallucination labels are directly aligned with the model's internal reasoning processes through a Visual-NLI framework.
\item We conduct comprehensive experiments on four major VQA datasets, addressing various hallucination types, and demonstrate FaithSCAN's effectiveness across multiple research questions.
\end{itemize}

The remainder of this paper is organized as follows.
Section~\ref{sec:related} reviews related work.
Section~\ref{sec:preliminaries} defines the problem setting and introduces key concepts.
Section~\ref{sec:method} presents the proposed methodology.
Section~\ref{sec:exp_setup} describes the experimental setup, including datasets, metrics, and implementation.
Section~\ref{sec:experiments} reports results and analysis.
Finally, Section~\ref{sec:conclusion} concludes the paper.

\section{Related Work}
\label{sec:related}
\subsection{Hallucination Detection in Visual Question Answering}
Faithfulness hallucination detection is critical in VQA, where generated answers can be fluent and confident yet inconsistent with visual evidence~\cite{chen2025survey}. A variety of detection methods have been proposed in recent years, which can be broadly categorized into \emph{external verification-based} and \emph{internal signal- or uncertainty-driven} approaches. External verification methods explicitly check cross-modal consistency between answers and visual inputs, often using object detectors, VQA models, scene graphs, or LLM evaluators~\cite{li-etal-2023-evaluating,jing2024faithscore,gu2024survey}. By decomposing answers into atomic claims and validating them against visual evidence, they enable fine-grained hallucination detection. However, these methods rely on multi-stage pipelines and pretrained components, incurring high computational cost and sensitivity to error propagation, while largely ignoring the VLM's internal reasoning process~\cite{jing2024faithscore}. Internal signal- or uncertainty-driven methods, in contrast, exploit the model’s outputs, attention patterns, and intermediate embeddings to detect hallucinations, offering a more direct and efficient view into the generation process~\cite{liu2025paying,jiang2025devils,ghosh2025visual}.

\subsection{Uncertainty-Based Hallucination Detection}

Model-internal uncertainty has emerged as an effective and relatively inexpensive cue for hallucination detection. Both language models and VLMs exhibit increased predictive uncertainty when hallucinations occur~\cite{zhang2023enhancing,quevedo2024detecting,aryal2025howard,gu2024survey}. Methods such as SE~\cite{farquhar2024detecting} and likelihood-based measures~\cite{ma2025semantic,zhao2025sese,jiang2025interpreting} quantify uncertainty by aggregating variations across multiple decoding outputs. In VQA, extensions incorporate visual perturbations or multimodal semantic clustering, e.g., vision-amplified semantic entropy~\cite{liao2025vision} and dense geometric entropy~\cite{gautam2025hedge}, which estimate hallucination likelihood by comparing semantic distributions under clean and perturbed visual inputs. These methods demonstrate that uncertainty signals can effectively capture hallucinations, including in safety-critical domains such as medical VQA~\cite{liao2025vision}. Nevertheless, most approaches rely on repeated sampling or perturbations to obtain reliable estimates, which increases inference cost and can be sensitive to stochastic decoding behavior.

\subsection{Sources and Exploitation of Internal Uncertainty Signals}

Hallucinations in VQA are not only reflected in the magnitude of predictive uncertainty but also in structured patterns across attention, intermediate representations, and logits. Early studies leveraged attention dynamics to detect hallucinations, e.g., DHCP~\cite{zhang2025dhcp} captures cross-modal attention patterns, showing consistent differences between hallucinated and faithful responses. OPERA~\cite{huang2024opera} demonstrates that hallucinations correlate with over-reliance on sparse summary tokens in self-attention, which can be mitigated through dynamic reweighting. Feature-based methods, such as HallE-Control~\cite{zhai2023halle}, reduce hallucinations by projecting intermediate embeddings to suppress reliance on unverifiable visual details. Embedding- or logit-based approaches, e.g., PROJECTAWAY~\cite{jiang2025interpreting} and ContextualLens~\cite{phukan2025beyond}, exploit token-level predictions and context-aware embeddings to localize and mitigate hallucinations at fine-grained object or patch levels. Despite their effectiveness, these methods are often limited by partial access to internal signals or simplified modeling of cross-modal interactions. Together, they indicate that hallucination-related cues are encoded across heterogeneous internal signals, highlighting the feasibility and necessity of fully exploring and exploiting them for reliable detection.

More recently, HaluNet~\cite{tong2025halunet} studies efficient hallucination detection in text-based question answering using internal uncertainty signals. Our work instead focuses on faithfulness hallucinations in VQA and extends this paradigm through explicit fusion of multimodal internal evidence. In addition, we provide a more systematic design and empirical validation of this methodology in the multimodal setting.

\section{Preliminaries}
\label{sec:preliminaries}
\subsection{Visual Question Answering}
VQA aims to generate a natural language answer conditioned on an input image and a textual question. Formally, let:
\begin{equation}
I \in \mathcal{I}, \quad q \in \mathcal{Q},
\end{equation}
denote the input image and question, respectively, where $\mathcal{I}$ and $\mathcal{Q}$ denote the spaces of images and textual questions.
Given $(I, q)$, a VLM parameterized by $\theta$ produces an answer:
\begin{equation}
\hat{a} = f_\theta(I, q), \quad \hat{a} \in \mathcal{A},
\end{equation}
where $\mathcal{A}$ denotes the space of textual responses.

In this work, we focus on modern VLMs that generate answers in an autoregressive manner. 
Specifically, all three VLM architectures studied in this paper (see Section~\ref{subsec:vlm_arch}) follow this paradigm, modeling the conditional distribution as:
\begin{equation}
p_\theta(a \mid I, q) = \prod_{t=1}^{T} p_\theta(a_t \mid a_{<t}, I, q),
\end{equation}
where $a$ denotes the output token sequence.

\subsection{Faithfulness Hallucination Detection in VQA}

We focus on \emph{faithfulness hallucinations}~\cite{huang2025survey}, where the generated answer is inconsistent with the visual evidence present in the input image. Given a VQA instance:
\begin{equation}
(I, q, \hat{a}),
\end{equation}
the goal of hallucination detection is to determine whether the answer $\hat{a}$ is visually grounded in $I$.

We formulate faithfulness hallucination detection as a supervised binary classification problem. Specifically, we aim to learn a classifier:
\begin{equation}
g_\phi : (I, q, \hat{a}) \rightarrow y,
\quad y \in \{0,1\},
\end{equation}
where $y = 1$ indicates a faithfulness hallucination and $y = 0$ denotes a faithful answer.

Unlike generation-based uncertainty estimation methods, our learning objective is not to approximate the full output distribution $p_\theta(a \mid I, q)$, but to directly model:
\begin{equation}
p_\phi(y \mid I, q, \hat{a}),
\end{equation}
using uncertainty signals and representations internally encoded in the VLM during a \emph{single forward pass}. This formulation enables efficient hallucination detection without repeated sampling, external verification models, or additional knowledge sources.

\begin{table*}[t]
\centering
\caption{Explicit mapping from observed hallucination phenomena to failure pathways in the multimodal reasoning process.}
\label{tab:phenomenon_to_pathway}
\renewcommand{\arraystretch}{1.2}
\begin{tabular}{p{3.8cm} p{5.5cm} p{6.2cm}}
\toprule
\textbf{Observed Hallucination Phenomenon} 
& \textbf{Typical Manifestation in VQA} 
& \textbf{Primary Failure Pathway} \\
\midrule
\textbf{Object Hallucination} 
& Mentioning non-existent objects or attributes not supported by the image 
& \textbf{Perceptual Failure}: inaccurate or incomplete visual entity representation \\
\midrule
\textbf{Relational Hallucination} 
& Incorrect spatial, logical, or attribute relations between visual entities 
& \textbf{Perceptual $\rightarrow$ Reasoning Failure}: unreliable visual grounding propagates to relational inference \\
\midrule
\textbf{Scene Hallucination} 
& Misinterpreting global scene context or high-level visual semantics 
& \textbf{Perceptual $\rightarrow$ Reasoning Failure}: failure in compositional reasoning and cross-modal semantic alignment \\
\midrule
\textbf{False-Premise Hallucination} 
& Answering questions with invalid or contradictory premises without rejection 
& \textbf{Interaction-Level Failure}: incorrect grounding of the question--answer interaction \\
\bottomrule
\end{tabular}
\end{table*}

\subsection{Vision--Language Model Architectures}
\label{subsec:vlm_arch}
Existing vision--language models vary in how they align visual and textual modalities. We focus on three representative models that illustrate distinct paradigms.
\begin{itemize}
    \item \textbf{InstructBLIP}~\cite{dai2023instructblip} uses query-based alignment, producing compact visual embeddings that interact with the language model via queries.
    \item \textbf{LLaVA-NeXT}~\cite{liu2023visual} relies on projection-based alignment, treating visual features as additional text tokens concatenated with language embeddings.
    \item \textbf{Qwen-VL}\cite{Qwen3-VL} employs a unified multimodal transformer, where visual and textual tokens interact throughout all layers.
\end{itemize}

Despite these architectural differences, all models encode rich uncertainty information, including token-level predictive uncertainty, visual reliability, and cross-modal alignment. This motivates hallucination detection methods that leverage internal representations rather than relying solely on generation-level sampling. Details of each architecture are provided in Supplementary Material Sec.~B.1.

\section{Methodology}
\label{sec:method}

\subsection{Method Overview}

\begin{figure*}[!htbp]
    \centering
    \includegraphics[width=\textwidth]{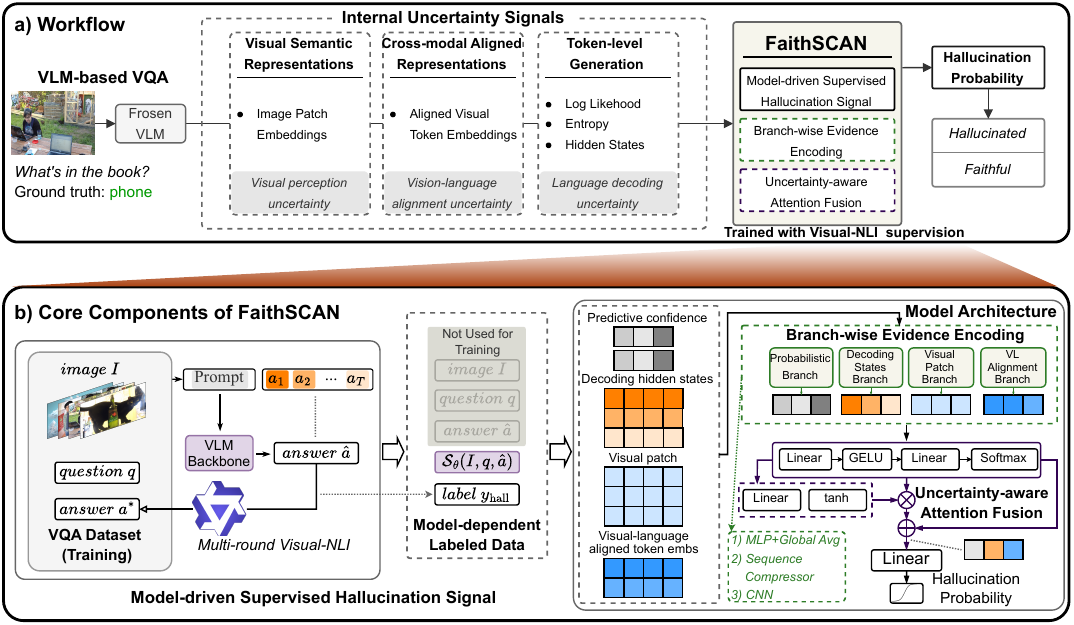}
    
\caption{Overview of FaithSCAN for single-pass hallucination detection in VQA. (a) Workflow of our method: multiple types of internal uncertainty signals are extracted from a frozen VLM in a single forward pass and combined with model-driven supervision to train FaithSCAN. (b) Core components of FaithSCAN, including the construction of supervision signals and the model architecture.}

    \label{fig:fig2}
\end{figure*}

In this work, we propose a \emph{single-pass, model-driven framework} for multimodal hallucination detection in VQA (Fig.~\ref{fig:fig2}). 
Unlike prior uncertainty-based methods that rely on repeated sampling or external verification, our approach operates entirely within a single forward pass of a VLM, leveraging its internal representations as supervisory signals. 
Hallucination detection is formulated as a \emph{supervised classification problem} informed by multiple complementary sources of uncertainty: (i) token-level predictive uncertainty during answer generation, (ii) intermediate visual representations encoding the model’s perception, and (iii) visually grounded representations aligned with the textual semantic space. 
Being fully \emph{model-driven}, the method does not require external estimators or auxiliary generative processes, making it compatible with diverse VLM architectures such as query-based (InstructBLIP), projection-based (LLaVA), and tightly coupled multimodal (Qwen3-VL) models. 
By explicitly learning the mapping between observed hallucination phenomena, their manifestations in VQA, failure pathways of multimodal reasoning, and internal model states (Table~\ref{tab:phenomenon_to_pathway}), our approach provides a principled and efficient solution for hallucination detection without expensive inference cost.

\subsection{Why Supervised Hallucination Detection}

Most uncertainty-based hallucination detection methods~\cite{farquhar2024detecting,zhang2024vl,ma2025semantic,tong2025semantic,zhao2025sese} aim to estimate or calibrate uncertainty, yet converting these scores into decisions still requires carefully chosen thresholds, limiting practical utility. 
By framing detection as a supervised decision-making problem, we directly map internal uncertainty signals to hallucination labels, producing actionable outputs and simplifying deployment. As described in Section~\ref{subsec:FaithSCAN}, this approach learns from internal signals rather than calibrating model uncertainty, reducing learning difficulty: learning focuses on extracting useful internal signals rather than output-space confidence modeling.

\subsection{Low-Cost Supervision via Model-Driven Labeling}
A central challenge in supervised hallucination detection is obtaining reliable supervision at scale. Manual annotation of multimodal hallucinations is costly, and many benchmarks provide only correct answers or predefined hallucination references~\cite{goyal2017vqa,li-etal-2023-evaluating,wang2024haloquest,park2025halloc}. Automatic annotation methods can alleviate this~\cite{chen2024mllm}, but they typically produce \emph{model-agnostic} labels that capture generic failures rather than the \emph{model-dependent} nature of hallucinations, which stem from a model’s generation behaviors, inductive biases, and cross-modal alignment deficiencies. Consequently, supervision decoupled from the target model is inherently limited.

We generate hallucination labels tightly coupled with the target model’s decoding dynamics~\cite{goyal2017vqa,li-etal-2023-evaluating,wang2024haloquest,park2025halloc}, derived from semantic consistency between generated and reference answers conditioned on the same image--question context. A strong external VLM serves as a semantic judge~\cite{gu2024survey,chen2024mllm}, but it is not a detached oracle; it evaluates answers in the context of the target model’s reasoning, producing model-aware labels. Visual grounding constraints enforced via a structured Visual-NLI protocol ensure that labels capture model-specific failure modes. We then verify label reliability through complementary automatic and human-based procedures, which is critical for supervised hallucination detection.

\subsubsection{Answer-Level Hallucination Labels Conditioned on Gold Answers}
\label{subsec:answer_level_labels}

We consider VQA datasets that provide a reference answer $a^\ast$ for each image--question pair $(I, q)$.
Given a VLM $f_\theta$, we first obtain its most likely answer $\hat{a}$ under a low-temperature decoding regime.
The objective is to determine whether $\hat{a}$ is faithful to the visual evidence and semantically consistent with the reference answer.

To this end, we formulate hallucination labeling as a \emph{visual natural language inference} (VNLI) problem.
Specifically, we construct the triplet $(I, q, a^\ast)$ as the premise and treat the model-generated answer $\hat{a}$ as the hypothesis.
A pretrained vision--language judge $\mathcal{J}$ is then queried to assess the semantic relationship between the premise and the hypothesis, conditioned on the image.

Formally, the judge outputs a categorical probability distribution:
\begin{equation}
\mathbf{p} = \mathcal{J}(I, q, a^\ast, \hat{a}) \in \Delta^{3},
\end{equation}
where $\mathbf{p} = (p_{\text{ent}}, p_{\text{con}}, p_{\text{unc}})$ corresponds to the probabilities of \emph{entailment}, \emph{contradiction}, and \emph{uncertainty}, respectively.
The three judgment categories are defined as follows:
\begin{itemize}
    \item \textbf{Entailment}: the hypothesis $\hat{a}$ is semantically consistent with the reference answer $a^\ast$, and any additional details are clearly supported by the image.
    \item \textbf{Contradiction}: the hypothesis conflicts with the reference answer or describes visual attributes that are inconsistent with the image.
    \item \textbf{Uncertainty}: the image does not provide sufficient evidence to verify the hypothesis, or the hypothesis introduces details that are unsupported but not directly contradicted.
\end{itemize}

To improve robustness against stochasticity in the judge model, we perform $R$ independent judging rounds and average the predicted probabilities:
\begin{equation}
\bar{\mathbf{p}} = \frac{1}{R} \sum_{r=1}^{R} \mathbf{p}^{(r)}.
\end{equation}
We then define the hallucination probability as the aggregated mass of non-entailment outcomes:
\begin{equation}
p_{\text{hall}} = \bar{p}_{\text{con}} + \bar{p}_{\text{unc}}.
\end{equation}
The final answer-level hallucination label is determined by:
\begin{equation}
y_{\text{hall}} = \mathbb{I}\!\left(p_{\text{hall}} > \bar{p}_{\text{ent}}\right),
\end{equation}
where $\mathbb{I}(\cdot)$ denotes the indicator function.

In this work, we treat both \emph{contradiction} and \emph{uncertainty} as indicators of hallucination, as they reflect failures of visual grounding or over-commitment beyond the available visual evidence.

We emphasize that the judge model is only used during dataset construction.
At inference time, the proposed hallucination detector operates independently, relying solely on the internal representations of the target VLM.

\subsubsection{Human Verification via Small-Scale Multi-Round Sampling}

While the proposed model-driven labeling strategy enables scalable supervision, it is still essential to assess the reliability of the automatically constructed hallucination labels.
To this end, we introduce a small-scale human verification protocol that serves as a quality control mechanism rather than a primary annotation source.

Concretely, we randomly sample a subset of VQA instances (100 samples per data set) and inspect the most-likely answer generated by the target VLM under low-temperature decoding.
For each sampled instance, human annotators are presented with the image, question, gold answer, and model-generated answer, and must determine whether the model answer constitutes a hallucination, defined as being unfaithful to the visual evidence or inconsistent with the reference answer.
To avoid confounding factors, samples with ambiguous questions or problematic ground-truth references are explicitly marked and excluded from subsequent agreement analysis.

Importantly, human annotation is not used to replace model-driven labels.
Instead, it is employed to validate whether the hallucination signals produced by the Visual-NLI judge are consistent with human judgment when applied to the same model outputs.
Let $y^{\text{human}} \in \{0,1\}$ denote the human label indicating non-hallucination or hallucination, and $y^{\text{model}} \in \{0,1\}$ denote the corresponding binary label derived from the model-driven NLI judgment.
We evaluate the agreement between $y^{\text{human}}$ and $y^{\text{model}}$ using multiple complementary metrics, including raw agreement ratio, Cohen’s $\kappa$, and Matthews correlation coefficient (MCC); detailed analyses and discussion are provided in Section~\ref{exp:rq2}.

\subsubsection{Automated Label Reliability Verification}
\label{subsec:reliability_weight}
Assessing the reliability of such labels through human verification remains costly and inherently limited.
We therefore introduce an \emph{automated label reliability verification} framework that aims to quantify the reliability of model-driven hallucination labels in a scalable manner, and to examine how such reliability estimates can be incorporated into the training process.

Given an image--question--answer triplet, the verification procedure evaluates the consistency of the assigned hallucination label across multiple model-based assessments, including repeated semantic verification~\cite{farquhar2024detecting}, stochastic answer generation~\cite{tong2025semantic}, and explicit self-reflection.
These assessments capture complementary aspects of decisiveness, robustness, and self-consistency, and are aggregated into an instance-level reliability score that reflects confidence in the supervision signal rather than the likelihood of hallucination itself.

We further investigate the use of this reliability score as an uncertainty-aware signal to modulate the contribution of individual training instances, without altering the original labels or introducing inference-time overhead.
Detailed definitions of the methods, strategies for training integration, and preliminary experimental validations are provided in Supplementary Material Sections~D.2, D.3.

\subsection{Uncertainty-Related Internal Signals in Vision--Language Models}
\label{subsec:uncertainty_signals}
In this subsection, we describe three categories of internal signals that are accessible in mainstream VLMs and serve as the foundation of our hallucination detection method.
Our approach is based on the observation that faithfulness hallucinations in VLMs are closely tied to the model’s internal uncertainty during inference.
Rather than introducing external supervision or repeated sampling, we directly extract uncertainty-related signals from intermediate representations produced in a single forward pass of the VLM $f_\theta$ (Fig.~\ref{fig:fig2}a).
Importantly, these signals are \emph{model-driven} and \emph{architecture-dependent}:
their concrete instantiation depends on the design of a given VLM, while their functional roles can be unified under a common formalization.

\subsubsection{Problem Setting}
Given an image--question pair $(I, q)$, the VLM generates an answer sequence $\hat{a} = (a_1, \ldots, a_T)$ autoregressively according to the conditional distribution $p_\theta(\hat{a} \mid I, q) = \prod_{t=1}^{T} p_\theta(a_t \mid a_{<t}, I, q)$.
During this generation process, the model produces a collection of intermediate representations spanning visual encoding, cross-modal alignment, and language decoding.
We extract uncertainty-related signals from these internal representations without modifying the model parameters $\theta$ or the inference procedure.

\subsubsection{Token-Level Generation Signals}
At each decoding step $t$, the language model outputs a predictive distribution:
\begin{equation}
p_\theta(\cdot \mid a_{<t}, I, q).
\end{equation}
From this distribution and the associated decoder hidden states, we extract three token-level signals: the log-likelihood of the generated token,
\begin{equation}
\ell_t = \log p_\theta(a_t \mid a_{<t}, I, q),
\end{equation}
the entropy of the predictive distribution,
\begin{equation}
H_t = - \sum_{v \in \mathcal{V}} p_\theta(v \mid a_{<t}, I, q)\log p_\theta(v \mid a_{<t}, I, q),
\end{equation}
and the hidden embedding $\mathbf{h}_t \in \mathbb{R}^{d_h}$ of the generated token from a selected decoder layer.
Here, $d_h$ denotes the dimensionality of the decoder hidden states.
While $\ell_t$ and $H_t$ provide scalar measures of predictive confidence, the hidden embedding $\mathbf{h}_t$ is treated as a high-dimensional representation that implicitly encodes semantic uncertainty.

\subsubsection{Visual Semantic Representations}
\label{subsec:visual_semantic_repr}

Prior to cross-modal interaction, the vision encoder extracts visual features from the input image $I$.
We denote the resulting set of patch-level visual embeddings as:
\begin{equation}
\mathbf{V}^{\mathrm{raw}} = \{ \mathbf{v}_i \}_{i=1}^{N}, \qquad \mathbf{v}_i \in \mathbb{R}^{d_v},
\end{equation}
where $N$ is the number of visual patches and $d_v$ denotes the dimensionality of the visual embedding space.
Each $\mathbf{v}_i$ corresponds to a localized image region.
In practice, $\mathbf{V}^{\mathrm{raw}}$ is obtained from the final hidden states of the vision backbone before any language conditioning or multimodal fusion.
This representation reflects the model’s internal visual perception state and serves as a latent visual semantic space that may encode uncertainty in visual understanding.

\subsubsection{Cross-Modal Aligned Representations}
\label{subsec:cross_modal_aligned_repr}

Modern VLMs employ an explicit alignment module to project visual representations into a language-compatible semantic space.
We denote the resulting aligned visual tokens as:
\begin{equation}
\mathbf{V}^{\mathrm{align}} = \{ \tilde{\mathbf{v}}_j \}_{j=1}^{M}, \qquad \tilde{\mathbf{v}}_j \in \mathbb{R}^{d_a},
\end{equation}
where $M$ is the number of aligned visual tokens and $d_a$ denotes the dimensionality of the aligned multimodal embedding space.
These representations are produced by a model-specific alignment function:
\begin{equation}
\mathbf{V}^{\mathrm{align}} = \mathcal{T}_\theta(\mathbf{V}^{\mathrm{raw}}, q),
\end{equation}
which may incorporate cross-attention, projection layers, or query-dependent pooling mechanisms.
The aligned visual tokens directly participate in the autoregressive decoding of $\hat{a}$ and therefore reflect visual information as mediated by vision--language interaction.

\subsubsection{Unified View of Uncertainty Signals}
Although the concrete extraction of the above signals depends on the architecture of the VLM, they can be jointly formalized as a set of internal uncertainty-related representations produced during the generation of $\hat{a}$, namely:
\begin{equation}
\mathcal{S}_\theta(I, q, \hat{a}) =
\Big(
\{ (\ell_t, H_t, \mathbf{h}_t) \}_{t=1}^{T},
\mathbf{V}^{\mathrm{raw}},
\mathbf{V}^{\mathrm{align}}
\Big),
\end{equation}
which collectively capture uncertainty arising from language decoding, visual semantic perception, and vision--language interaction, and constitute a model-internal, single-pass characterization of uncertainty.
In the next subsection, we describe how these heterogeneous signals are jointly modeled to produce a hallucination confidence score.

\subsection{Model Architecture}
\label{subsec:FaithSCAN}

We propose FaithSCAN, a lightweight supervised hallucination detection network that operates on internal uncertainty signals extracted from a frozen VLM, as illustrated in Fig.~\ref{fig:fig2}b.
FaithSCAN aggregates heterogeneous uncertainty evidence at the decision level and outputs a scalar hallucination score in a single forward pass.
Notably, the detector is \emph{model-dependent}: the dimensionality, number, and extraction locations of its inputs are determined by the architecture of the underlying VLM, enabling adaptation to model-specific uncertainty patterns without modifying the base model.

\subsubsection{Input Formulation}
Given the frozen VLM $f_\theta$, an image--question pair $(I,Q)$ and the model-generated answer $\hat{a}$, we extract $K$ internal uncertainty signals:
\begin{equation}
\mathcal{E} = \{ \mathbf{X}_1, \mathbf{X}_2, \dots, \mathbf{X}_K \},
\qquad \mathbf{X}_k \in \mathbb{R}^{L_k \times D_k}.
\end{equation}
Each $\mathbf{X}_k$ is a sequence-valued feature (e.g., token-level statistics, hidden states, or vision--language alignment features) obtained from different components of $f_\theta$.
These sources may vary in modality, sequence length, and feature dimension, and are treated independently without assuming a shared feature space.

\subsubsection{Branch-wise Evidence Encoding}
\label{subsec:branch_encoding}

Each uncertainty source $\mathbf{X}_k$ is processed by an independent branch encoder $f_k(\cdot)$ that compresses a variable-length sequence into a fixed-dimensional embedding:
\begin{equation}
\mathbf{h}_k = f_k(\mathbf{X}_k), 
\qquad \mathbf{h}_k \in \mathbb{R}^{d},
\end{equation}
where $d=64$ in all experiments.

All branch encoders are implemented as lightweight sequence compression modules rather than deep semantic encoders.
Specifically, we consider linear projections with global average pooling, trainable sequence compressors (Linear + LayerNorm + ReLU + mean pooling), and convolutional encoders with adaptive pooling~\cite{kim2014convolutional}, 
where one-dimensional convolutions are applied with kernel size $3$ and stride $1$. For a detailed description of these branch encoders, refer to Supplementary Material Sec.~B.5.
Despite architectural differences, all variants share the same objective: to summarize the reliability and stability of internal signals while limiting representational capacity to reduce overfitting.
We further analyze the generalization capability under distribution shifts in Section~\ref{sec:OOD}, while the effectiveness of architectural variants is examined in Section~\ref{sec:token_uncertainty_construction}.

\subsubsection{Uncertainty-aware Attention Fusion}
\label{subsec:uncertainty_fusion}
Let $\mathbf{H} = [\mathbf{h}_1, \ldots, \mathbf{h}_K] \in \mathbb{R}^{K \times d}$ denote the set of evidence embeddings produced by the evidence encoders.
To integrate heterogeneous evidence representations, we adopt a cross-branch attention mechanism:
\begin{equation}
\alpha_k =
\frac{
\exp\!\left(\mathbf{w}_a^\top \mathrm{GELU}\!\left(\mathbf{W}_a \mathbf{h}_k\right)\right)
}{
\sum_{j=1}^{K}
\exp\!\left(\mathbf{w}_a^\top \mathrm{GELU}\!\left(\mathbf{W}_a \mathbf{h}_j\right)\right)
},
\qquad
\mathbf{h} = \sum_{k=1}^{K} \alpha_k \mathbf{h}_k .
\end{equation}

This attention computes adaptive weights over branches, allowing the model to emphasize certain evidence representations while aggregating information across all branches.

To further modulate the fused representation, we apply a gated residual transformation:
\begin{equation}
\mathbf{g} = \tanh\!\left(\mathbf{W}_g \mathbf{h}\right),
\qquad
\widetilde{\mathbf{h}} = \mathbf{h} \odot \mathbf{g} + \mathbf{h},
\end{equation}
where the $\tanh(\cdot)$ gate modulates each feature of the fused representation, and the residual connection preserves the original information while allowing adaptive adjustment.

\subsubsection{Hallucination Prediction}

The fused representation $\widetilde{\mathbf{h}}$ is mapped to a hallucination probability via:
\begin{equation}
p = \sigma\!\left(\mathbf{w}^\top \widetilde{\mathbf{h}} + b\right),
\end{equation}
where $p \in (0,1)$ denotes the likelihood that the generated answer is hallucinated.
FaithSCAN is trained end-to-end with model-driven supervision while keeping the underlying VLM frozen, using a binary cross-entropy loss over the hallucination labels.

\begin{table*}[t]
\centering
\caption{Datasets used in our experiments and their coverage of hallucination types.}
\label{tab:dataset}
\begin{tabular}{lcccccc}
\toprule
\textbf{Dataset} & \textbf{\#Samples} & \textbf{Split} & \textbf{Ground-Truth Verification} &
\textbf{Object} & \textbf{Relation / Scene} & \textbf{False Premise} \\
\midrule
HalLoc-VQA   & 80,540 / 10,932 / 10,082\textsuperscript{\dag} & Train/Val/Test  & Token-level verification & \checkmark & \checkmark & -- \\
POPE         & 9,000 & Test & Object-level verification & \checkmark & -- & -- \\
HaloQuest    & 7,140 / 608 & Train/Test & Answer-level verification & \checkmark & Partial & \checkmark \\
VQA v2       & 443,757 / 214,354 / 447,793 & Train/Val/Test & Answer-level verification & \checkmark & \checkmark & -- \\
\bottomrule
\end{tabular}
\begin{flushleft}
$^{\dag}$ HalLoc overall contains about 155k labeled samples across tasks including VQA; the VQA subset used in our work is drawn from this set.
\end{flushleft}
\end{table*}


\section{Experimental Setup}
\label{sec:exp_setup}
\subsection{Datasets}

We evaluate our approach on four multimodal datasets with complementary characteristics: \textbf{HalLoc-VQA}~\cite{park2025halloc}, \textbf{POPE}~\cite{li-etal-2023-evaluating}, \textbf{HaloQuest}~\cite{wang2024haloquest}, and \textbf{VQA v2}~\cite{goyal2017vqa}, which differ in hallucination types, annotation granularity, and inducing mechanisms. HalLoc-VQA provides token-level supervision for object- and scene-level hallucinations. POPE focuses on object-level hallucinations via direct probing of object existence, HaloQuest emphasizes false-premise and reasoning-induced hallucinations, and VQA v2 serves as a naturalistic evaluation set for real-world visual grounding challenges. Dataset statistics are summarized in Table~\ref{tab:dataset}; for training and evaluation, we sample 2,000 training instances and 1,000 evaluation instances from each split (full HaloQuest test set used) and report averages over three random seeds, ensuring consistent evaluation while controlling computational cost, following common practice in multimodal hallucination and uncertainty analysis~\cite{lee2024vhelm,chen-etal-2025-unveiling,park2025glsim,farquhar2024detecting}.

\subsection{Baselines}
We compare FaithSCAN with a diverse set of representative baselines for multimodal hallucination detection, focusing on uncertainty-based methods at token, latent, and semantic levels. 
Token-level baselines include \textbf{Predictive Entropy (PE)}~\cite{holtzman2020curious} and \textbf{Token-level Negative Log-Likelihood (T-NLL)}~\cite{zhang2023enhancing}, measuring per-token predictive uncertainty. 
Latent-space baselines such as \textbf{Embedding Variance (EmbVar)}~\cite{grewal2024improving} and \textbf{Semantic Embedding Uncertainty (SEU)}~\cite{grewal2024improving} capture instability in vision-conditioned answer embeddings. 
Semantic-level baselines include \textbf{Semantic Entropy (SE)}~\cite{farquhar2024detecting} and a multimodal adaptation of \textbf{SelfCheckGPT}~\cite{manakul2023selfcheckgpt}, detecting contradictions across answer pairs. 
A multimodal \textbf{$P(\text{True})$} verification baseline~\cite{farquhar2024detecting} evaluates answer correctness via image-conditioned yes/no prompts. 
Finally, a \textbf{Logistic Regression}~\cite{hosmer2013applied} classifier trained on our dataset serves as a supervised reference, with implementation details and key hyperparameters provided in Supplementary Material Sec.~C.

\subsection{Implementation Details and Metrics}

We evaluate FaithSCAN on three vision--language backbones:
\textbf{IB-T5-XL}, \textbf{LLaVA-8B}, and \textbf{Qwen3-VL-8B}\footnote{
Full model names:
IB-T5-XL (Salesforce/instructblip-flan-t5-xl);
LLaVA-8B (llava-hf/llama3-llava-next-8b-hf);
Qwen3-VL-8B (Qwen/Qwen3-VL-8B-Instruct).
}.
Uncertainty signals, including token-level log-likelihood, predictive entropy, and hidden representations from a selected transformer layer,
are extracted in a single forward pass and fused via an attention-based module after projection to 64 dimensions.
Hallucination supervision is obtained via model-driven verification using \textbf{Qwen2.5-VL-32B-Instruct} with low-temperature sampling ($T=0.1$).
Models are trained with AdamW and binary cross-entropy loss with logits for up to 40 epochs,
with early stopping based on validation AUROC.
Prompt templates are provided in Supplementary Material Sec.~A, and model implementation details are given in Sec.~B.

Focusing on the I2T setting where evaluation is performed primarily on generated textual answers, we assess hallucination detection using AUROC (ROC) to measure overall ranking quality, AURAC (RAC) to evaluate selective prediction under rejection, F1@Best (F1@B) to report the optimal classification trade-off, and RejAcc@50\% (RA@50) to quantify accuracy after rejecting the most uncertain half of predictions~\cite{farquhar2024detecting,zhao2025sese,tong2025semantic}. 
Together, these metrics provide a comprehensive and complementary evaluation of hallucination detection performance from both ranking and decision-making perspectives.

\begin{table*}[t]
\centering
\scriptsize
\caption{\textbf{RQ1:} Overall ID performance on three datasets and three backbone models.
Latency is reported as the average sampling and inference time per 100 examples (in seconds), with colors indicating increasing computational cost from green to red.
In the \emph{Sample} column, red marks the slowest methods requiring multiple samplings per query.
\textbf{Bold} and \underline{underlined} values denote the best and second-best results, respectively.}
\label{tab:main-id-table}
\resizebox{\textwidth}{!}{%
\begin{tabular}{cl|cccccccccccccccc}
\toprule
\multicolumn{1}{l}{} &   & 
\multicolumn{4}{c}{\textbf{HalLoc-VQA}} & 
\multicolumn{4}{c}{\textbf{HaloQuest}} & 
\multicolumn{4}{c}{\textbf{VQA v2}} & 
\multicolumn{2}{c}{\textbf{Latency (s/100 samples)}} \\
\cmidrule(lr){3-6} \cmidrule(lr){7-10} \cmidrule(lr){11-14} \cmidrule(lr){15-16}
\multicolumn{1}{l}{\multirow{-2}{*}{}} &
  \multirow{-2}{*}{} &
  ROC & RAC & F1@B & RA@50 &
  ROC & RAC & F1@B & RA@50 &
  ROC & RAC & F1@B & RA@50 &
  Sample & Infer \\ 
\midrule
 & PE & 0.536 & 0.526 & 0.730 & 0.527 & 0.578 & 0.526 & 0.805 & 0.534 & 0.624 & 0.239 & 0.561 & 0.225 & \cellcolor{red!10} 1326.308 & \cellcolor{green!10}0.002 \\
 & T-NLL & 0.535 & 0.537 & \textbf{0.762} & 0.542 & 0.571 & 0.550 & 0.730 & 0.549 & 0.623 & 0.265 & 0.502 & 0.285 & \cellcolor{red!10}1257.512 & \cellcolor{green!10}0.001 \\
 & EmbVar & 0.657 & 0.464 & 0.746 & 0.453 & 0.605 & 0.521 & 0.790 & 0.578 & 0.724 & 0.773 & \textbf{0.667} & 0.845 & \cellcolor{red!20}1430.884 & \cellcolor{green!10}0.060 \\
 & SE & 0.583 & 0.563 & 0.653 & 0.562 & 0.537 & 0.595 & 0.805 & 0.608 & 0.565 & 0.311 & 0.491 & 0.297 & \cellcolor{red!20}1381.929 & \cellcolor{red!10}4.913 \\
 & SEU & 0.663 & 0.452 & 0.738 & 0.440 & 0.660 & 0.498 & 0.776 & 0.515 & \textbf{0.772} & \underline{0.787} & \underline{0.667} & \textbf{0.868} & \cellcolor{red!20}1685.975 & \cellcolor{yellow!30}1.845 \\
 & SelfCheckGPT & 0.509 & 0.545 & 0.722 & 0.590 & 0.474 & 0.623 & 0.797 & 0.637 & 0.502 & 0.684 & 0.514 & 0.657 & \cellcolor{red!20}1602.887 & \cellcolor{red!10}78.613 \\
 & $P(\text{True})$ & \underline{0.675} & 0.597 & 0.753 & 0.565 & 0.515 & 0.609 & 0.770 & 0.642 & 0.703 & 0.782 & 0.554 & 0.777 & \cellcolor{red!10}1127.961 & \cellcolor{red!10}14.938 \\
 & Logistic & 0.669 & \underline{0.669} & 0.734 & \underline{0.674} & \textbf{0.735} & \underline{0.833} & \underline{0.855} & \underline{0.807} & 0.710 & 0.760 & 0.555 & 0.806 & \cellcolor{red!10}1570.349 & \cellcolor{green!10}0.087 \\
\multirow{-9}{*}{\rotatebox{90}{IB-T5-XL}} & FaithSCAN (Ours) & \textbf{0.754} & \textbf{0.778} & \underline{0.755} & \textbf{0.792} & \underline{0.713} & \textbf{0.854} & \textbf{0.870} & \textbf{0.885} & \underline{0.752} & \textbf{0.820} & 0.573 & \underline{0.858} & \cellcolor{red!10}1560.237 & \cellcolor{green!10}0.605 \\
   \midrule
 & PE & 0.605 & 0.276 & 0.535 & 0.258 & 0.440 & 0.368 & 0.515 & 0.348 & 0.553 & 0.588 & 0.433 & 0.565 & \cellcolor{red!10}1491.448 & \cellcolor{green!10}0.002 \\
 & T-NLL & 0.607 & 0.256 & 0.529 & 0.247 & 0.447 & 0.372 & 0.446 & 0.373 & 0.558 & 0.725 & 0.413 & 0.787 & \cellcolor{red!10}1524.740 & \cellcolor{green!10}0.001 \\
 & EmbVar & 0.747 & \underline{0.820} & 0.509 & 0.855 & 0.546 & 0.355 & 0.515 & 0.284 & 0.593 & 0.437 & 0.481 & 0.345 & \cellcolor{red!20}2504.584 & \cellcolor{green!10}0.175 \\
 & SE & 0.683 & 0.747 & 0.462 & 0.795 & 0.595 & 0.280 & 0.557 & 0.275 & 0.630 & 0.186 & 0.425 & 0.172 & \cellcolor{red!20}2632.833 & \cellcolor{red!10}11.212 \\
 & SEU & \underline{0.769} & 0.799 & \underline{0.667} & \underline{0.865} & 0.536 & 0.324 & 0.434 & 0.328 & 0.604 & 0.666 & 0.429 & 0.645 & \cellcolor{red!20}2443.890 & \cellcolor{yellow!30}1.769 \\
 & SelfCheckGPT & 0.715 & 0.779 & 0.625 & 0.818 & 0.603 & 0.576 & 0.510 & 0.534 & \textbf{0.693} & 0.741 & \underline{0.533} & 0.755 & \cellcolor{red!20}2588.092 & \cellcolor{red!10}340.683 \\
 & $P(\text{True})$ & 0.615 & 0.690 & 0.520 & 0.748 & 0.716 & 0.761 & 0.553 & 0.730 & 0.663 & 0.772 & \textbf{0.564} & \underline{0.825} & \cellcolor{red!10}1178.084 & \cellcolor{red!10}19.511 \\
 & Logistic & 0.733 & 0.788 & 0.580 & 0.830 & \textbf{0.834} & \underline{0.840} & \underline{0.746} & \underline{0.857} & 0.592 & \underline{0.797} & 0.384 & 0.802 & \cellcolor{red!10}1874.895 & \cellcolor{green!10}0.114 \\
\multirow{-9}{*}{\rotatebox{90}{LLaVA-8B}} & FaithSCAN (Ours) & \textbf{0.779} & \textbf{0.823} & \textbf{0.680} & \textbf{0.873} & \underline{0.827} & \textbf{0.848} & \textbf{0.762} & \textbf{0.902} & \underline{0.685} & \textbf{0.801} & 0.494 & \textbf{0.831} & \cellcolor{red!10}1898.950 & \cellcolor{green!10}0.741 \\
   \midrule
 & PE & 0.686 & 0.178 & 0.535 & 0.200 & 0.573 & 0.726 & 0.229 & 0.789 & 0.549 & 0.394 & 0.428 & 0.305 & \cellcolor{red!10}1328.434 & \cellcolor{green!10}0.002 \\
 & T-NLL & 0.650 & 0.233 & 0.524 & 0.190 & 0.542 & 0.733 & 0.257 & 0.784 & 0.524 & 0.193 & 0.312 & 0.195 & \cellcolor{red!10}1330.996 & \cellcolor{green!10}0.001 \\
 & EmbVar & 0.628 & 0.705 & \underline{0.570} & 0.787 & 0.641 & 0.875 & 0.444 & 0.902 & 0.587 & 0.777 & \underline{0.483} & 0.823 & \cellcolor{red!20}2194.586 & \cellcolor{green!10}0.107 \\
 & SE & 0.559 & 0.325 & 0.456 & 0.305 & 0.606 & 0.873 & 0.316 & 0.873 & 0.532 & 0.208 & 0.374 & 0.205 & \cellcolor{red!20}2019.877 & \cellcolor{red!10}13.812 \\
 & SEU & 0.653 & 0.742 & \textbf{0.587} & 0.800 & 0.678 & 0.868 & 0.500 & \underline{0.917} & 0.614 & 0.805 & 0.396 & 0.848 & \cellcolor{red!20}2199.427 & \cellcolor{yellow!30}1.821 \\
 & SelfCheckGPT & 0.656 & 0.716 & 0.552 & 0.748 & 0.783 & 0.878 & \underline{0.501} & 0.903 & \underline{0.712} & 0.851 & 0.430 & 0.877 & \cellcolor{red!20}2165.013 & \cellcolor{red!10}367.522 \\
 & $P(\text{True})$ & \textbf{0.750} & \textbf{0.825} & 0.548 & \textbf{0.838} & 0.631 & 0.776 & 0.394 & 0.832 & \textbf{0.777} & \underline{0.860} & \textbf{0.499} & \textbf{0.915} & \cellcolor{red!10}1143.720 & \cellcolor{red!10}24.299 \\
 & Logistic & 0.672 & 0.686 & 0.496 & 0.724 & \underline{0.795} & \underline{0.887} & 0.446 & 0.910 & 0.675 & 0.812 & 0.411 & 0.810 & \cellcolor{red!10}1689.458 & \cellcolor{green!10} 0.102 \\
\multirow{-9}{*}{\rotatebox{90}{Qwen3-VL-8B}} & FaithSCAN (Ours) & \underline{0.739} & \underline{0.757} & 0.569 & \underline{0.811} & \textbf{0.829} & \textbf{0.941} & \textbf{0.532} & \textbf{0.975} & 0.695 & \textbf{0.876} & 0.420 & \underline{0.884} & \cellcolor{red!10}1741.652 & \cellcolor{yellow!30} 1.191 \\
\bottomrule
\end{tabular}
}
\end{table*}

\section{Experiments}
\label{sec:experiments}
In this section, we conduct extensive experiments to evaluate the proposed method from multiple perspectives. Specifically, our experimental study is organized around four research questions.
\textbf{RQ1 (Overall Effectiveness and Generalization):} How effective is the proposed method in detecting hallucinations across different vision--language models and datasets covering diverse hallucination types?
\textbf{RQ2 (Quality of Model-driven Supervision Signals):} How accurate and dependable are the model-driven supervision signals, and to what extent can they support hallucination detection in a supervised learning setup?
\textbf{RQ3 (Internal Representation Analysis):} What roles do different internal representations, including token-level uncertainty, visual embeddings, and aligned visual--semantic embeddings, play in hallucination detection, and how do they complement each other?
\textbf{RQ4 (Practical Utility and Interpretability):} How does the proposed method perform under practical decision-making scenarios, and to what extent are its decisions interpretable?

\subsection{Overall Effectiveness and Generalization (RQ1)}

We evaluate FaithSCAN under in-distribution (ID) and out-of-distribution (OOD) settings across four datasets. FaithSCAN leverages token-level uncertainty signals (log-likelihood, entropy, embeddings) and visual patch features. 
Sequence-level representations are obtained via global average pooling and fused through attention (see Section~\ref{sec:internal_representation_analysis} for architectural choices).

\subsubsection{In-distribution Evaluation}
ID results (Table~\ref{tab:main-id-table}) show that FaithSCAN reaches or surpasses SOTA performance, achieving up to 8\% AUROC improvement and maintaining top performance across metrics and backbone models. Sampling-based baselines such as SEU, SelfCheckGPT, and $P(\text{True})$ perform competitively but incur higher sampling and inference costs, whereas FaithSCAN achieves strong detection in a single pass. SE underperforms due to long-form, detailed answers that undermine clustering assumptions. Performance further improves with more capable backbones, suggesting richer internal representations provide clearer uncertainty signals.

\begin{table}[t]
\centering
\caption{\textbf{RQ1:} OOD hallucination detection performance (AUROC) under cross-dataset and multi-source training.
In cross-dataset evaluation, models are trained on a single dataset and tested on unseen datasets, while multi-source results are obtained by joint training on all available datasets.
L and F denote logistic regression and FaithSCAN (ours), respectively.
AUROC values below 0.5 are highlighted in red, indicating generalization failure, and for multi-source training, the better result between L and H is highlighted in green.}
\label{tab:ood_result}
\resizebox{\columnwidth}{!}{%
\begin{tabular}{l cc cc cc}
\toprule
\multirow{2}{*}{\textbf{Train $\rightarrow$ Test}}
& \multicolumn{2}{c}{\textbf{IB-T5-XL}}
& \multicolumn{2}{c}{\textbf{LLaVA-8B}}
& \multicolumn{2}{c}{\textbf{Qwen3-VL-8B}} \\
\cmidrule(lr){2-3} \cmidrule(lr){4-5} \cmidrule(lr){6-7}

& L & F
& L & F
& L & F \\

\midrule
\multicolumn{7}{l}{\emph{Single-source training (cross-dataset evaluation)}} \\

HalLoc-VQA $\rightarrow$ POPE
& 0.499& 0.603 & 0.671& 0.703 & 0.559& 0.632 \\

HalLoc-VQA $\rightarrow$ HaloQuest
& 0.514& 0.508 & 0.601& \cellcolor{red!20}0.393 & \cellcolor{red!20}0.493& \cellcolor{red!20}0.412 \\

HalLoc-VQA $\rightarrow$ VQA v2
& 0.631& 0.669 & 0.546& 0.571 & 0.578& 0.624 \\

\midrule

HaloQuest $\rightarrow$ HalLoc-VQA
& 0.555& 0.567 & 0.501& 0.506 & 0.521& \cellcolor{red!20}0.429 \\

HaloQuest $\rightarrow$ POPE
& 0.641& 0.573 & 0.580& 0.511 & 0.631& 0.639 \\

HaloQuest $\rightarrow$ VQA v2
& 0.569& 0.561 & 0.567& 0.585 & 0.579& 0.594 \\

\midrule

VQA v2 $\rightarrow$ HalLoc-VQA
& 0.628& 0.640 & 0.598& 0.571 & 0.556& 0.605 \\

VQA v2 $\rightarrow$ POPE
& 0.629& 0.712 & 0.628& 0.604 & 0.644& 0.701 \\

VQA v2 $\rightarrow$ HaloQuest
& 0.541& 0.664 & 0.700& 0.759 & 0.650& 0.717 \\

\midrule
\multicolumn{7}{l}{\emph{Multi-source training (trained on all datasets)}} \\

All $\rightarrow$ HalLoc-VQA
& 0.703& \cellcolor{green!20}0.743& 0.663& \cellcolor{green!20}0.700& 0.611& \cellcolor{green!20}0.672\\

All $\rightarrow$ HaloQuest
& 0.696& \cellcolor{green!20}0.697& 0.777& \cellcolor{green!20}0.796& \cellcolor{green!20}0.757& 0.749\\

All $\rightarrow$ VQA v2
& 0.709& \cellcolor{green!20}0.715& 0.631& \cellcolor{green!20}0.638& 0.618& \cellcolor{green!20}0.668\\

All $\rightarrow$ POPE
& 0.680& \cellcolor{green!20}0.744& 0.669& \cellcolor{green!20}0.675& 0.654& \cellcolor{green!20}0.689\\

\bottomrule
\end{tabular}
}
\end{table}

Moreover, we evaluated the computational efficiency of each method in both the sampling and inference stages. In the sampling phase (e.g., obtaining answers or extracting embeddings), most existing methods are relatively slow. Regarding inference, approaches that rely on multiple samples or use external models exhibit far lower efficiency than those that infer directly from the model’s internal states. Because FaithSCAN integrates multiple embeddings, its inference speed is slightly reduced when handling larger embedding sizes; nevertheless, it remains among the fastest, highlighting the advantages of our approach.

\subsubsection{Out-of-distribution Evaluation}
\label{sec:OOD}

OOD evaluation (Table~\ref{tab:ood_result}) reveals limited generalization for models trained on a single dataset, especially when hallucination types differ (e.g., object/scene-level vs. false-premise). FaithSCAN exhibits stronger transfer than logistic regression, and multi-source training enhances OOD performance, highlighting the importance of diverse training data. Nonetheless, a performance gap between ID and OOD underscores the challenge of hallucination detection under distribution shifts, motivating future work on improving generalization of model-driven uncertainty approaches.

\begin{figure}[t]
    \centering
    \includegraphics[width=\columnwidth]{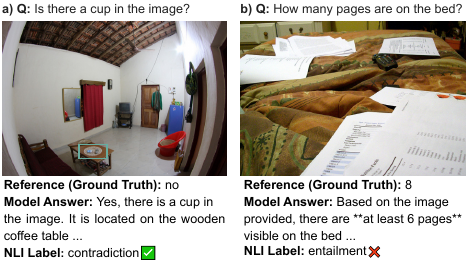}
    
    \caption{\textbf{RQ2:} Illustrative examples of the consistency judgment process based on model-driven supervision, showing both positive and negative cases. (a) The sample is drawn from POPE, where the LLM-based judger correctly identifies a \emph{contradiction} relationship between the model-generated answer and the ground-truth reference. (b) The sample is drawn from VQA v2, where the generated answer is not strictly precise; nevertheless, it is judged as \emph{entailment} by the LLM judge.}

    \label{fig:RQ2_example}
\end{figure}

\subsection{Quality of Model-driven Supervision Signals (RQ2)}
\label{exp:rq2}
\subsubsection{Human Evaluation}
We evaluate whether the proposed model-driven supervision reliably approximates human annotations for hallucination detection.

Qualitative examples are shown in Fig.~\ref{fig:RQ2_example}. In Fig.~\ref{fig:RQ2_example}a, a POPE sample demonstrates correct detection of a \emph{contradiction} by the LLM-based judger, reflecting the dataset’s clear object-level hallucination boundaries. Fig.~\ref{fig:RQ2_example}b illustrates a failure case from VQA v2, where minor factual errors are masked by semantic similarity, leading to an incorrect \emph{entailment} judgment.

\begin{table}[t]
\centering
\caption{\textbf{RQ2:} Consistency analysis between model-driven supervision signals and human annotations across different datasets.}
\label{tab:model_driven_agreement}
\resizebox{\columnwidth}{!}{%
\begin{tabular}{lccc}
\toprule
\textbf{Dataset} & \textbf{Agreement} $\uparrow$ & \textbf{Cohen's Kappa} $\uparrow$ & \textbf{MCC} $\uparrow$ \\
\midrule
HalLoc-VQA & 0.960 & 0.834 & 0.837 \\
POPE       & 0.990 & 0.947 & 0.948 \\
HaloQuest  & 0.960 & 0.858 & 0.861 \\
VQA v2     & 0.950 & 0.798 & 0.799 \\
\bottomrule
\end{tabular}
}
\end{table}

Quantitatively (Table~\ref{tab:model_driven_agreement}), the proposed supervision exhibits strong agreement with human annotations, achieving agreement scores in the range of 0.950–0.990. Moreover, Cohen’s Kappa and MCC are consistently high across datasets, with most values exceeding 0.8.

These results indicate that the model-driven supervision effectively aligns with human judgment, providing a reliable signal for hallucination detection across diverse scenarios.

\subsubsection{Automated Evaluation}
\label{subsec:ae}

\begin{figure}[t]
\centering
\includegraphics[width=\columnwidth]{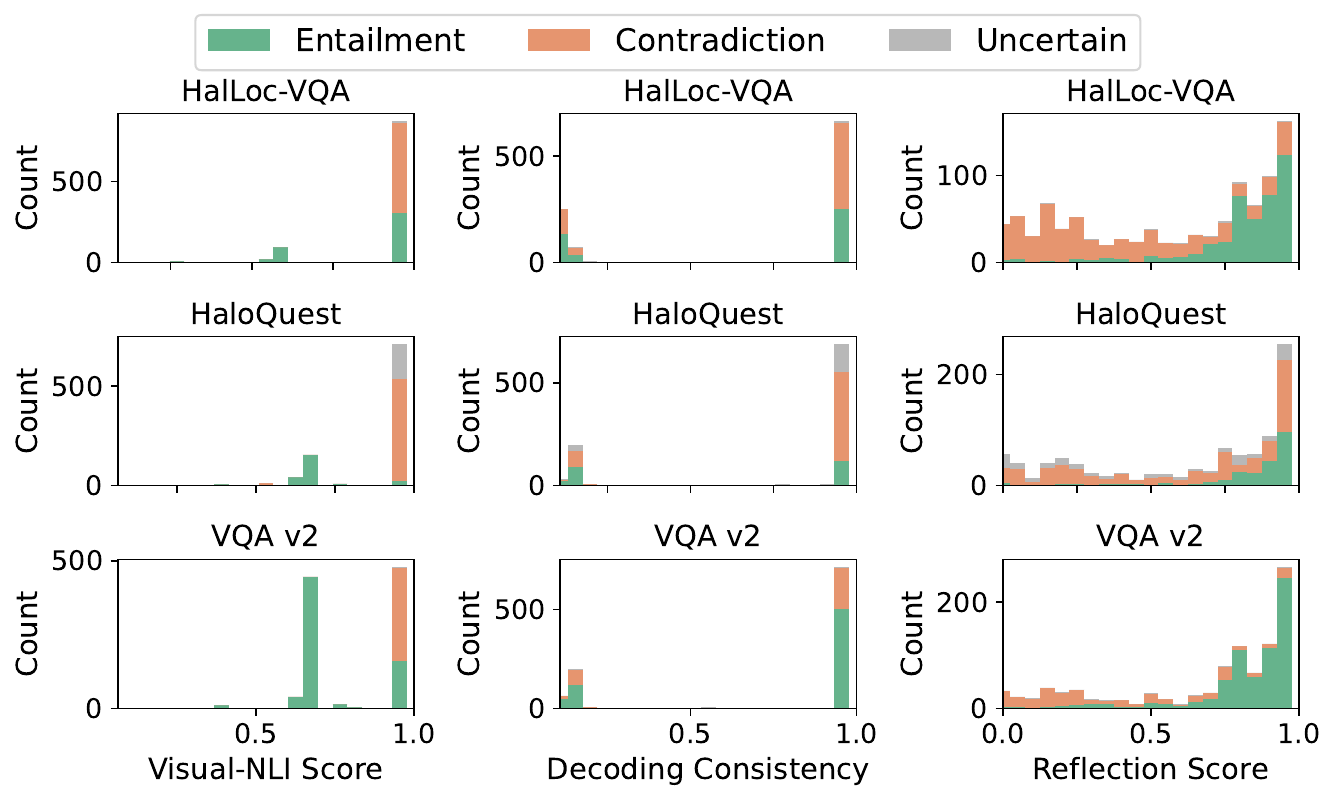}
\caption{\textbf{RQ2:} Distributions of post-hoc reliability scores on three datasets. Answers are from IB-T5-XL; bars are colored by the original three-way hallucination labels (Sec.~\ref{subsec:answer_level_labels}). Results for other backbones are in Supplementary Material~Sec.~D.2.}

\label{fig:automated_eval}
\end{figure}

We evaluate the reliability of hallucination labels for answers generated by different VLM backbones using three post-hoc scores: Visual-NLI decisiveness ($s_\text{nli}$), stochastic decoding consistency ($s_\text{stoch}$), and reflection-based self-consistency ($s_\text{ref}$). Their distributions are shown in Fig.~\ref{fig:automated_eval}.

$s_\text{nli}$ is high for most samples, indicating that the external model produces stable judgments across NLI rounds, with slightly lower consistency on the more challenging VQA v2. $s_\text{stoch}$ shows a similar trend across all three datasets, suggesting the generation model maintains consistent internal confidence across different data distributions. $s_\text{ref}$ provides a finer-grained view. Hallucinated samples tend to have lower $s_\text{ref}$ because unfaithful outputs make reflection less stable. Differences across datasets also appear: HaloQuest contains some uncertain cases due to false-premise hallucinations, while hallucinations in VQA v2 and HalLoc-VQA are more often clearly judged as entailment or contradiction.

These scores show that model-driven hallucination labels are generally stable, while reflection provides deeper insight into the link between data distributions and hallucination labels, highlighting a promising direction for future study.

\begin{figure*}[!htbp]
    \centering
    \includegraphics[width=\textwidth]{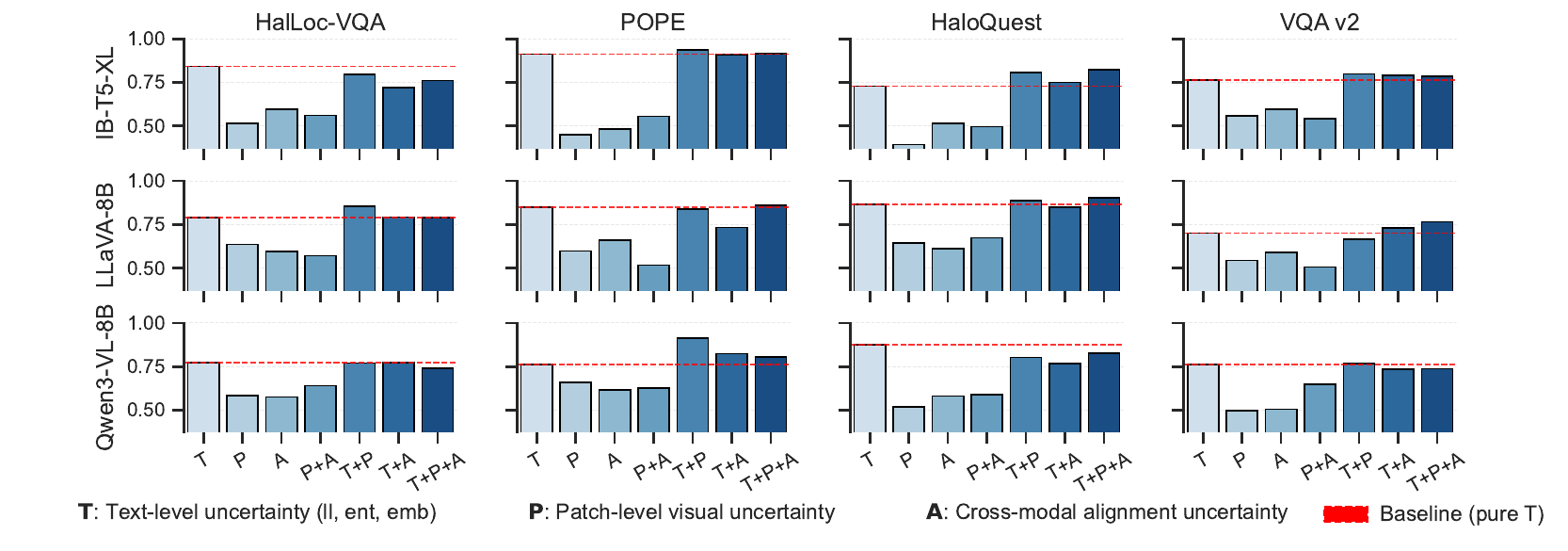}
    
    \caption{
    \textbf{RQ3:} AUROC scores for different uncertainty source combinations used in hallucination detection across the training sets of four datasets and three base models. Labels indicate uncertainty types: T = text-level (log-likelihood, entropy, embedding), P = patch-level visual, and A = cross-modal alignment.
    }
    \label{fig:inde_ab}
\end{figure*}

\begin{figure}[!htb]
    \centering

    \begin{subfigure}[t]{\columnwidth}
        \centering
        \includegraphics[width=\columnwidth]{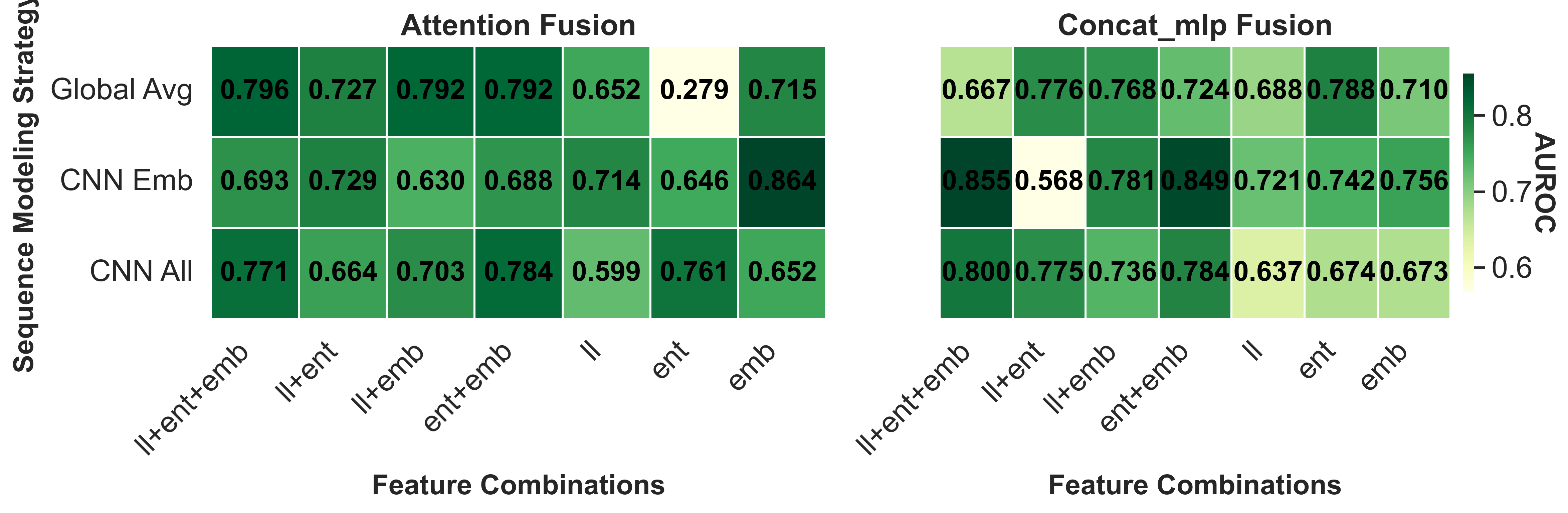}
        \caption{Results of IB-T5-XL.}
        \label{fig_rq3_1:subfig1}
    \end{subfigure}

    \vspace{0.5em} 

    \begin{subfigure}[t]{\columnwidth}
        \centering
        \includegraphics[width=\columnwidth]{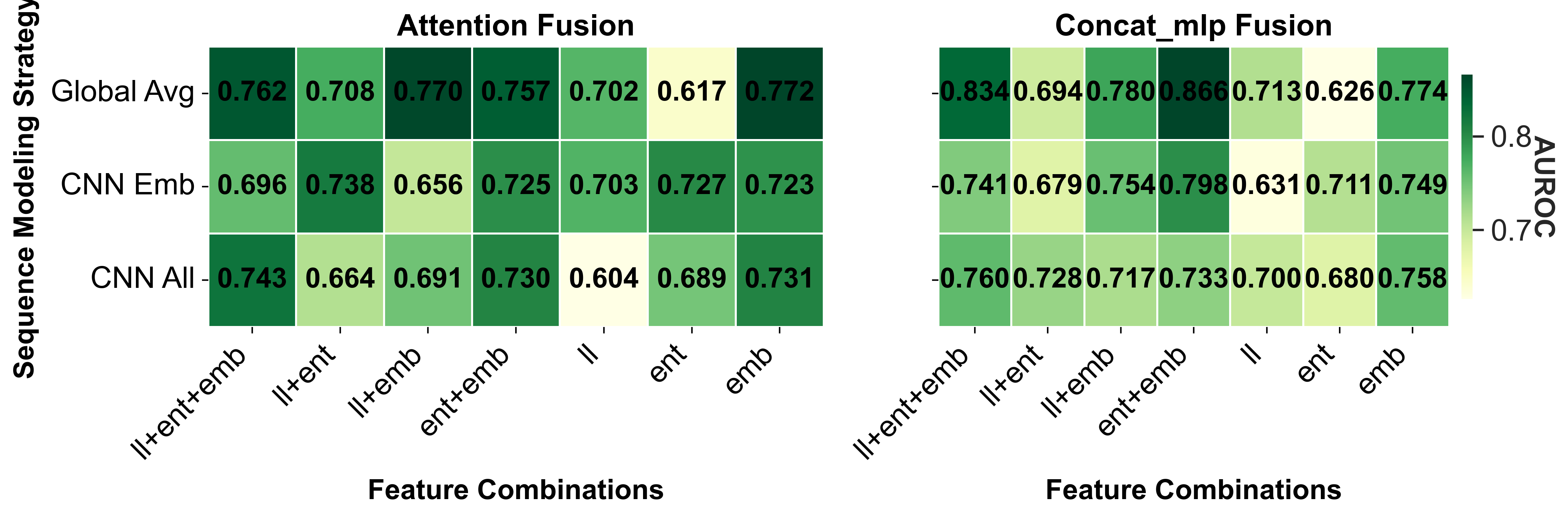}
        \caption{Results of LLaVA-8B.}
        \label{fig_rq3_1:subfig2}
    \end{subfigure}
    
        \vspace{0.5em} 

    \begin{subfigure}[t]{\columnwidth}
        \centering
        \includegraphics[width=\columnwidth]{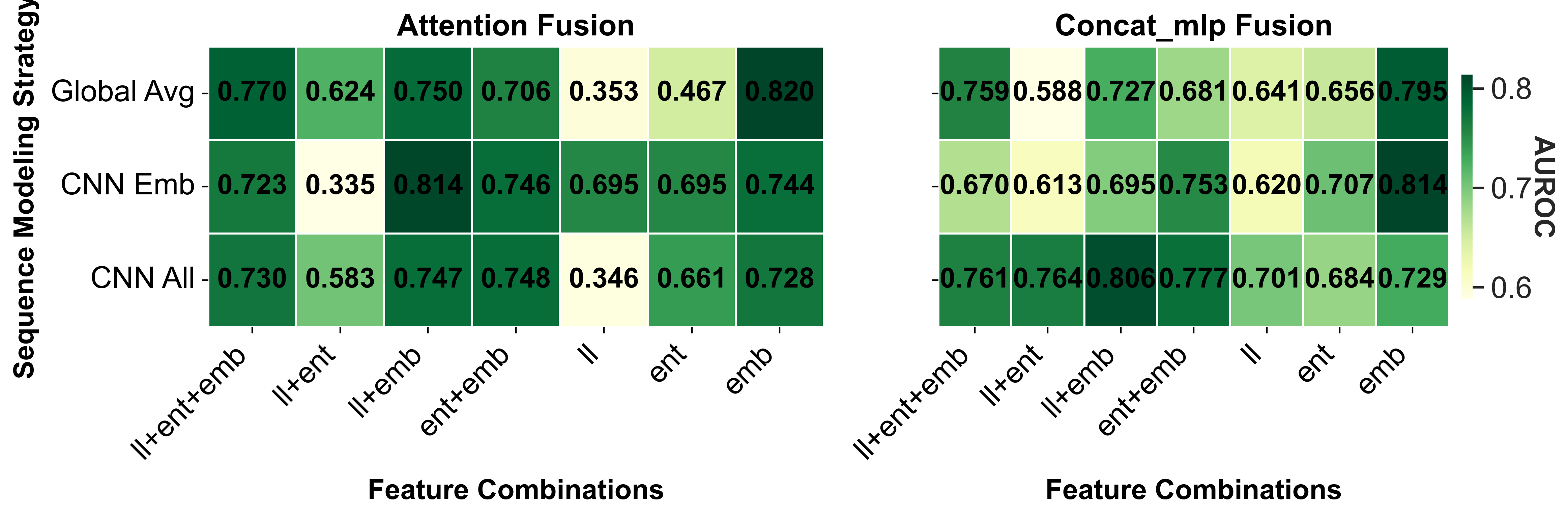}
        \caption{Results of Qwen3-VL-8B.}
        \label{fig_rq3_1:subfig3}
    \end{subfigure}

    \caption{\textbf{RQ3:} Ablation study on feature combination, sequence modeling strategy, and fusion methods on the HalLoc-VQA training set. The fusion methods include: Global Avg (average of all token-level features), CNN Emb (CNN modeling only on the embeddings), and CNN All (CNN modeling applied to all features).}
    \label{fig:combined}
\end{figure}

\subsection{Internal Representation Analysis (RQ3)}
\label{sec:internal_representation_analysis}

We investigate why FaithSCAN effectively detects hallucinations by analyzing the role of different internal representations and uncertainty sources. Experiments are conducted on HalLoc-VQA, POPE, HaloQuest, and VQA v2 using IB-T5-XL, LLaVA-8B, and Qwen3-8B.

\subsubsection{Contribution of Independent Uncertainty Sources}
\label{sec:independent_uncertainty_sources}

We investigate the contributions of independent uncertainty sources derived from distinct internal representations, including token-level uncertainty from language generation, visual uncertainty from compressed patch embeddings, and visual--semantic uncertainty from features aligned to the textual semantic space.

We first evaluate each source in isolation. As shown in Fig.~\ref{fig:inde_ab}, visual patch–level and visual--semantic uncertainty alone generally yield limited discriminative power across datasets, indicating that hallucinations are difficult to detect without access to token-level generation signals. In contrast, token-level uncertainty consistently provides a strong baseline, suggesting that many hallucination phenomena are rooted in the internal states of the language generation process.

The role of visual uncertainty, however, varies substantially across datasets and hallucination types. On POPE, which predominantly focuses on object-level hallucinations, integrating pure visual patch uncertainty leads to significant performance gains. This result suggests that object hallucinations are closely tied to perceptual failures, and that uncertainty at the visual perception level constitutes a critical failure pathway in such cases.

In contrast, on HalLoc-VQA and VQA v2, incorporating visual features yields limited or inconsistent improvements. These datasets involve more complex reasoning and compositional question answering, where hallucinations are less attributable to visual ambiguity and more strongly associated with failures in language modeling and multimodal reasoning. This observation reflects a shift in the dominant failure pathway from perception to higher-level reasoning processes.

A similar pattern is observed on HaloQuest, which primarily targets false-premise hallucinations. For models with relatively weaker visual capabilities, incorporating visual uncertainty substantially improves detection performance, indicating that interaction-level hallucinations can originate from unreliable visual grounding. However, for models with stronger visual encoders, the gains diminish, suggesting that hallucinations in this setting are more likely driven by reasoning-level failures. This further indicates that the source of hallucination is architecture-dependent and cannot be attributed to a single modality.

Overall, these results empirically support our methodological premise: effective hallucination detection requires explicitly modeling the mapping between observed hallucination phenomena, their manifestations in VQA, the underlying multimodal failure pathways, and the corresponding internal model states.

\subsubsection{Ablation Study on Token-level Uncertainty Modeling}
\label{sec:token_uncertainty_construction}

We evaluate token-level uncertainty signals derived from log-likelihoods, entropies, and hidden embeddings. As shown in Fig.~\ref{fig:combined}, embeddings provide the richest signal, followed by entropy, while log-likelihood contributes marginally. Combining these signals consistently improves performance across datasets and models, with attention-based fusion outperforming simple concatenation. Incorporating CNN-based sequence modeling on embeddings further enhances detection, highlighting that localized anomalies in embeddings capture hallucinations effectively. Supplementary Material Sec.~D.1 presents the ablation study on the attention implementation. Based on these results, we adopt the combination of log-likelihood, entropy, embeddings, global average pooling, and attention as a stable FaithSCAN variant, which is used in all subsequent experiments.

\begin{figure}[t]
    \centering
    \includegraphics[width=\columnwidth]{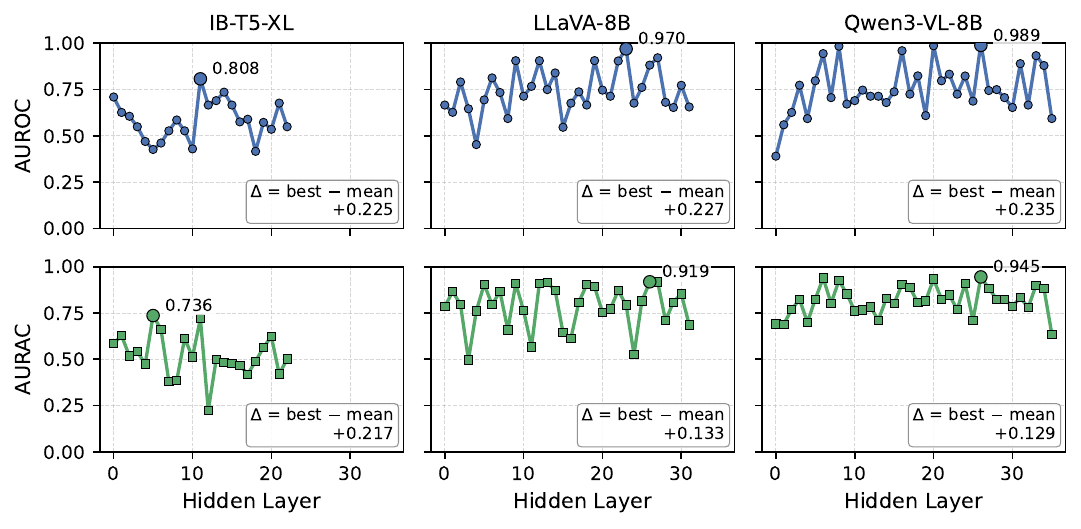}
    
    \caption{
    \textbf{RQ3:} Layer-wise contribution analysis of hallucination detection performance. The best-performing layer is highlighted, and $\Delta$ vs Mean denotes the relative improvement of the optimal layer over the average performance.
    }
    \label{fig:layer_ab}
\end{figure}

\subsubsection{Layer-wise Representation Analysis}
\label{sec:layer_wise_analysis}

We evaluate token-level representations from different decoder layers for hallucination detection on HalLoc-VQA using IB-T5-XL, LLaVA-8B, and Qwen3-8B. Each layer is tested on 100 samples, repeated three times, using all uncertainty signals from Section~\ref{subsec:uncertainty_signals} with attention-based fusion.
Fig.~\ref{fig:layer_ab} shows that representations from the earliest and deepest layers consistently underperform, while intermediate layers achieve higher performance across all models. The layer-wise uncertainty exhibits fluctuations, indicating that different layers encode complementary information: intermediate layers balance token-level variability and semantic abstraction, capturing cross-modal cues relevant to hallucination. This trend is particularly clear for Qwen3-8B and LLaVA-8B.

Based on these results, we select intermediate layers as token-level features in subsequent experiments. For most models we use the 2/3 depth layer, while for InstructBLIP we choose the 1/2 depth layer. This choice is guided by the layer-wise performance observed in this analysis and balances predictive performance with computational efficiency.

\begin{figure}[!htbp]
    \centering
    \includegraphics[width=\columnwidth]{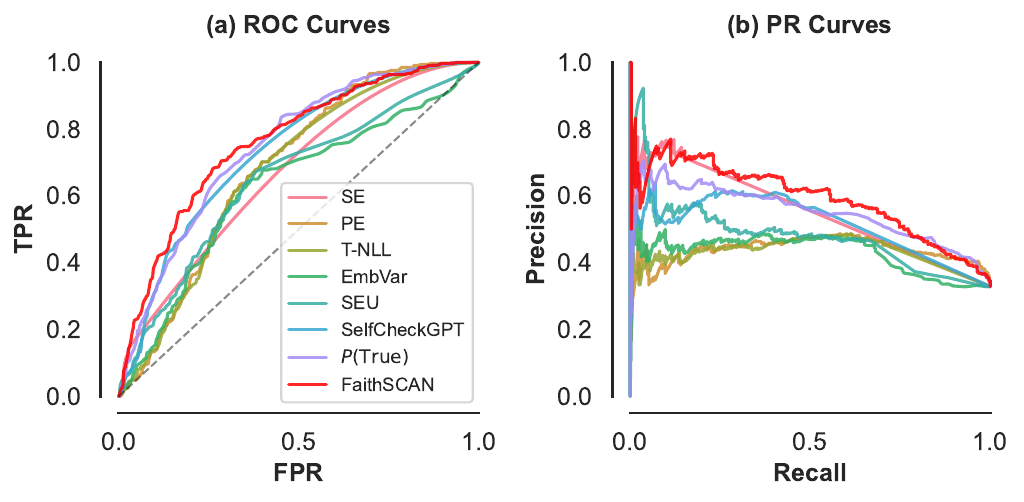}
    \caption{\textbf{RQ4:} ROC and PR curves of FaithSCAN and baseline methods on the HalLoc-VQA $\times$ Qwen3-VL-8B setting.}
    \label{fig:rq4-1}
\end{figure}

\subsection{Practical Utility and Interpretability (RQ4)}
\label{sec:rq4}

Hallucination detectors are typically deployed by applying a threshold on uncertainty or confidence scores, making performance sensitive to the precision–recall trade-off. We evaluate FaithSCAN under the HalLoc-VQA $\times$ Qwen3-VL-8B setting, reporting ROC and PR curves alongside baselines, and visualize normalized uncertainty distributions to illustrate separation between hallucinated and non-hallucinated samples.

To assess interpretability, we analyze token-level features using a gradient-based approach inspired by saliency analysis~\cite{simonyan2014deep,shrikumar2017learning}. For input features $\mathbf{x} = {x_{t,f}}$ and predicted probability $\hat{y} = \sigma(f_\theta(\mathbf{x}))$, attributions are computed as $A_{t,f} = (\partial \hat{y} / \partial x_{t,f}) \cdot x_{t,f}$, capturing both a feature’s influence on the prediction and its signal strength, yielding interpretable token-level contributions.

\begin{figure}[t]
    \centering
    \includegraphics[width=\columnwidth]{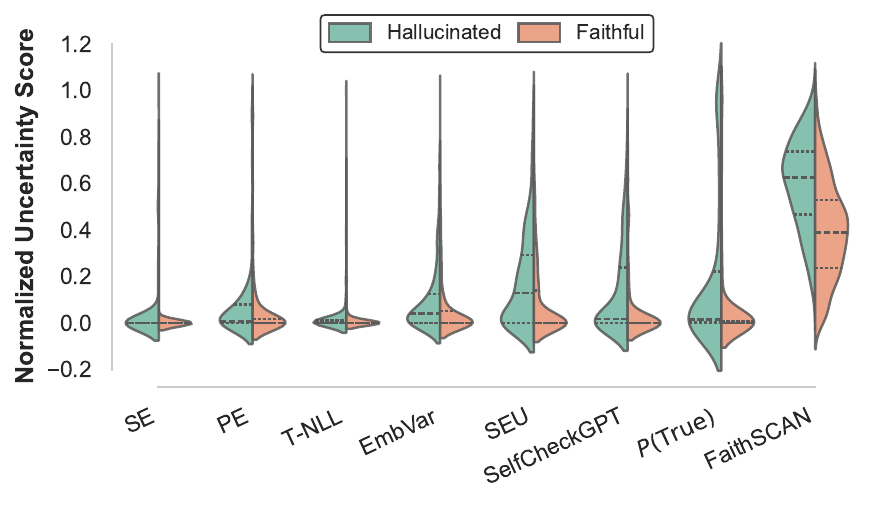}
    \caption{\textbf{RQ4:} Violin plots of normalized uncertainty scores for hallucinated and non-hallucinated samples.}
    \label{fig:rq4-2}
\end{figure}

\begin{figure}[t]
    \centering
    \includegraphics[width=\columnwidth]{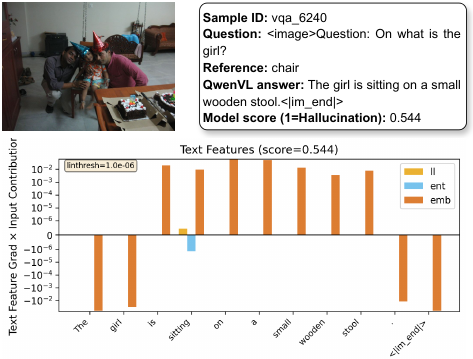}
    \caption{\textbf{RQ4:} Token-level feature attribution visualization using the \texttt{grad $\times$ input} method. Positive contributions highlight hallucination-related tokens, while negative contributions correspond to non-hallucinated regions.}
    \label{fig:rq4-3}
\end{figure}

Fig.~\ref{fig:rq4-1} presents the ROC and PR curves of FaithSCAN and competing baselines. Across most operating points, FaithSCAN consistently achieves superior performance, maintaining higher precision under the same recall levels.

To further examine the uncertainty modeling behavior, Fig.~\ref{fig:rq4-2} visualizes the normalized uncertainty score distributions using violin plots. Compared to baseline methods, FaithSCAN produces a markedly larger separation between hallucinated and non-hallucinated samples. Notably, although FaithSCAN is trained using binary supervision, it implicitly learns a meaningful uncertainty score that correlates well with hallucination likelihood.

Finally, Fig.~\ref{fig:rq4-3} presents a qualitative case study using token-level attributions from the \texttt{grad $\times$ input} method (see Supplementary Material Sec.~E). Fused visual features are difficult to map back to the original image in an orderly way, so we focus on text-side feature channels. Potential future analyses could leverage image perturbation techniques, as in~\cite{zhang2024vl}. In the example, Qwen3-VL-8B incorrectly identifies a \emph{stool} instead of a \emph{chair}, as it overlooks the key distinction of the backrest despite rich visual details. FaithSCAN assigns strong positive contributions to the hallucinated token sequence, accurately capturing the erroneous reasoning process. These highlighted regions align with human intuition, demonstrating the interpretability and diagnostic value of our approach.

\section{Conclusion}
\label{sec:conclusion}

This paper investigates faithfulness hallucination detection in VQA from a model internal perspective. Our results indicate that many hallucinations are already encoded in the internal states of VLMs during generation, rather than emerging solely at the output level. Uncertainty signals from language decoding, visual perception, and cross-modal interaction provide complementary views of different hallucination failure pathways. Their relative contributions vary across model architectures and hallucination types.

More broadly, this work suggests that hallucinations in VQA are structured phenomena rooted in multimodal reasoning dynamics. The effectiveness of lightweight and trainable detectors operating on frozen model states demonstrates that reliable hallucination detection can be achieved without repeated sampling or external verification. We hope these findings motivate future research to treat hallucination as an intrinsic property of vision–language reasoning, supporting the development of more reliable and interpretable multimodal systems.

\bibliographystyle{IEEEtran}
\bibliography{my,VQA}

\clearpage
\section*{Supplementary Material}
\addcontentsline{toc}{section}{Supplementary Material}

\setcounter{section}{0}

\section*{A. Prompt Templates and Input Formatting}

This section presents the prompt templates and input formatting strategies used in our framework.
To improve readability and reproducibility, all prompts are shown verbatim using a unified block-based format.

\subsection*{A.1 Visual-NLI Prompt for Model-Driven Label Generation}
\label{subsec:A1}
To generate model-driven supervision, we employ a Visual Natural Language Inference (Visual-NLI) prompt that evaluates whether a model-generated answer is supported by the image and consistent with the ground-truth answer.

Given an image, a question, a reference answer, and a hypothesis answer, the vision-language model is instructed to act as a strict Visual-NLI judge and output both a categorical label and a probability distribution.

\begin{promptbox}
\begin{lstlisting}
You are a strict Visual-NLI judge. Your goal is to evaluate whether the
HYPOTHESIS answer is supported by the IMAGE and consistent with the
QUESTION-REFERENCE pair.

### TASK
The REFERENCE is the ground-truth answer to the question.
Your job is to judge the HYPOTHESIS relative to the REFERENCE and the IMAGE.

### IMPORTANT RULES
1. You MUST verify every factual claim in the HYPOTHESIS against the IMAGE.
2. ENTAILMENT means:
   - Hypothesis semantically agrees with the reference, AND
   - Any additional details in the hypothesis must be clearly supported by the image.
3. CONTRADICTION means:
   - Hypothesis conflicts with the reference, OR
   - Hypothesis describes details that are clearly inconsistent with the image.
4. UNCERTAIN means:
   - Image does not provide enough visual evidence to verify the hypothesis, OR
   - Hypothesis contains extra details not contradicted but also not supported by the image.

### OUTPUT FORMAT
Return strict JSON only:
{"label": "...", "prob": {"entailment": p1, "contradiction": p2, "uncertain": p3}}

### INPUT
IMAGE: <image>
QUESTION: {question}
REFERENCE: {ground-truth answer}
HYPOTHESIS: {model output}
\end{lstlisting}
\end{promptbox}

The predicted probabilities are later used as soft supervision and reliability signals.

\subsection*{A.2 Reflection Prompt for Sample Reliability Estimation}

To estimate sample reliability, we design a reflection-based prompt that evaluates whether a \emph{fixed} model-generated answer can be consistently justified using only visual evidence from the image.

Unlike the Visual-NLI prompt used for label generation, this reflection prompt:
(i) does not introduce a reference answer,
(ii) does not assess factual correctness,
and (iii) explicitly forbids answer revision.
Instead, it focuses solely on the stability and sufficiency of visual grounding.

\begin{promptbox}
\begin{lstlisting}
You are given an image, a question, and a FIXED answer.

Question:
{question}

Answer (DO NOT change or rephrase this answer):
{answer}

Task:
Evaluate whether this answer can be consistently justified
using ONLY visual evidence from the image.

Guidelines:
- Do NOT propose a new or corrected answer.
- Do NOT judge correctness against any external reference.
- Focus on whether the answer relies on clearly identifiable
  visual evidence, rather than assumptions or hallucinated details.

Output ONLY a JSON object in the following format:
{
  "support_score": <number between 0 and 1>
}

A higher score means the answer has stronger and more stable
visual support.
\end{lstlisting}
\end{promptbox}

The resulting \texttt{support\_score} is used as a continuous reliability signal to weight samples during training.

\subsection*{A.3 Multimodal Prompt Structure for Model Inference}

For model inference, inputs are formatted as a multimodal user message consisting of an image and a textual question.
The image is provided through the visual channel, while the question is passed as plain text.

Conceptually, the prompt structure is as follows:

\begin{promptbox}
\begin{lstlisting}
[Image]
{question}
\end{lstlisting}
\end{promptbox}

The above structure is converted into a model-specific chat template using the official processor of each vision-language model, with a generation prompt appended automatically.
This abstraction ensures consistent multimodal conditioning while allowing model-specific tokenization and formatting.

\subsection*{A.4 Few-Shot Prompt for the P(True) Baseline}

For the $P(\text{True})$~\cite{farquhar2024detecting} baseline, we adopt a few-shot self-evaluation strategy in which the model is asked to judge whether a given answer to a question is correct.
Each prompt consists of two parts:
(i) a few-shot context composed of question--answer pairs with known correctness, and
(ii) a test instance for which the model predicts whether the answer is correct.

\paragraph*{Few-shot context}
Each few-shot example follows a fixed format, ending with an explicit correctness query and a positive confirmation.
Only samples that are accurate and non-hallucinatory are selected to construct the few-shot context.

\begin{promptbox}
\begin{lstlisting}
Answer the following question as briefly as possible.
Question: On which side of the photo is the man?
Answer: The man is on the right side of the photo.

Is this correct?
Yes

Question: Based on the image, respond to this question with a short answer:
On which side is the lady, the left or the right?
Answer: right

Is this correct?
Yes
\end{lstlisting}
\end{promptbox}

\paragraph*{Test instance}
After the few-shot context, a test question--answer pair is appended using the same format.
Unlike the few-shot examples, the final correctness judgment is left unanswered and must be predicted by the model.

\begin{promptbox}
\begin{lstlisting}
[Image]
Question: {question}
Answer: {candidate answer}

Is this correct? Yes or No
\end{lstlisting}
\end{promptbox}

The complete $P(\text{True})$ prompt is formed by concatenating the few-shot context and the test instance.
The probability assigned to the answer ``Yes'' is taken as the $P(\text{True})$ score.

In the multimodal setting, images are provided through the visual channel and are not serialized into text.

\subsection*{A.5 Prompt for the SelfCheckGPT Baseline}

For the SelfCheckGPT~\cite{manakul2023selfcheckgpt} baseline, we design a prompt that measures the degree of semantic contradiction between two answers to the same question.

Given a question, a candidate answer to be checked, and a reference answer, the model is instructed to output a scalar contradiction score between 0 and 1, without any explanation.

\begin{promptbox}
\begin{lstlisting}
You are given a question and two answers to it.
Your task is to estimate how contradictory the first answer (ri)
is with respect to the second answer (Sn).

First, consider the semantic relationship between the two answers,
including object identity, category membership, attributes, actions,
quantities, and temporal states.

Then assign a contradiction score based on the following scale:

- 0.0--0.1: The two answers are semantically equivalent or clear paraphrases.
- 0.2--0.3: Very minor differences (e.g., level of detail or synonyms).
- 0.4--0.6: Partial mismatch or underspecification.
- 0.7--0.8: Major semantic differences.
- 0.9--1.0: Completely contradictory or mutually exclusive answers.

Question: {question}
Answer to check (ri): {candidate answer}
Reference answer (Sn): {reference answer}

Output a single float between 0 and 1.
Do not explain your reasoning.
\end{lstlisting}
\end{promptbox}

This baseline operates purely at the answer level and does not explicitly incorporate visual evidence.

\section*{B. Model Implementation Details}

\subsection*{B.1 Detailed Vision--Language Model Architectures}
Existing VLMs differ in architectural design, but most share three functional components: a vision encoder, a language model, and a cross-modal alignment mechanism. We consider three representative models that cover distinct alignment paradigms.

\subsubsection{InstructBLIP}
InstructBLIP~\cite{dai2023instructblip} employs a \emph{query-based cross-modal alignment} strategy. A frozen vision encoder first extracts dense visual features from the image. These features are then queried by a lightweight Query Transformer (Q-Former), producing a compact set of visual query embeddings. The embeddings are projected into the input space of a Flan-T5 language model, enabling instruction-following VQA without fully fine-tuning the vision backbone or language model.

This design emphasizes compact visual abstraction and controlled vision--language interaction, but limits fine-grained token-level coupling between visual and textual modalities.

\subsubsection{LLaVA-NeXT}
LLaVA-NeXT adopts the same \emph{projection-based multimodal alignment} paradigm as the original LLaVA~\cite{liu2023visual}. Its vision encoder (\emph{vision tower}) extracts high-dimensional visual features, which are projected by a learnable multimodal layer and concatenated with text token embeddings. The combined embeddings are fed into a decoder-only LLaMA model, enabling cross-modal reasoning while treating visual features as additional tokens aligned to the language space. We use LLaVA-NeXT in this work as a stronger instantiation of the LLaVA architecture.

\subsubsection{Qwen-VL}
Qwen3-VL\cite{Qwen3-VL} adopts a \emph{unified multimodal transformer} architecture. Visual features are embedded and merged into a sequence of multimodal tokens, which are processed alongside textual tokens across multiple transformer layers. Unlike projection-based models, visual and textual representations interact throughout the network, allowing richer semantic alignment. This tight integration improves cross-modal reasoning but also complicates uncertainty propagation, which is critical for hallucination detection.

\subsection*{B.2 Extraction of Uncertainty Signals}
For each model, uncertainty-related signals are extracted in a single forward pass during autoregressive decoding along the most likely generation path (greedy decoding or very low temperature). Specifically, we collect:
\begin{itemize}
    \item \textbf{token-level log-likelihoods} (\texttt{ll}) and \textbf{predictive entropy} (\texttt{ent}) from the output distribution;
    \item \textbf{token-level hidden representations} (\texttt{emb}) from  a designated layer of the language model;
    \item \textbf{raw visual patch embeddings} (\texttt{mm\_patch}), capturing pre-fusion visual semantics;
    \item \textbf{cross-modal aligned visual embeddings} (\texttt{mm\_align}), representing image features after alignment with the language space.
\end{itemize}

Visual features are extracted via forward hooks at architecturally meaningful stages of each model. 
We distinguish between \emph{raw visual embeddings}, which encode purely visual information prior to multimodal fusion, and \emph{aligned visual embeddings}, which represent visual tokens projected into the language-aligned space.

For \textbf{IB-T5-XL (InstructBLIP)}, raw visual embeddings are taken from the vision encoder output after the final transformer layer and post-layer normalization, capturing patch-level visual representations before any interaction with the Q-Former or the language model. Aligned visual embeddings are obtained from the Q-Former cross-attention outputs, corresponding to image-conditioned visual query tokens.

For \textbf{LLaVA-8B}, raw visual embeddings are extracted from the post-layernorm outputs of the CLIP vision encoder, with the CLS token removed when present. Aligned visual embeddings are obtained from the multimodal projector, which maps visual tokens into the language embedding space and directly conditions text generation.

For \textbf{Qwen3-VL-8B}, raw visual embeddings are extracted from the output of the final visual transformer block before the patch merger, representing high-level visual tokens produced by the vision backbone. Aligned visual embeddings are obtained from the visual patch merger, which aggregates and projects visual tokens into the language embedding space for cross-modal interaction.

All visual embeddings are extracted once per input image and shared across all generated tokens, while token-level uncertainty signals are computed step-wise during autoregressive decoding.

\subsection*{B.3 Hallucination Labels}

Hallucination labels are obtained via a stronger external verifier:
\begin{itemize}
    \item Verifier model: \textbf{Qwen2.5-VL-32B-Instruct}.
    \item Sampling strategy: deterministic low-temperature decoding (temperature = 0.1, top-p = 1.0).
    \item Prompts and labels: a Visual-NLI style verification prompt that evaluates whether the generated answer is supported by the image and consistent with the reference, producing entailment/contradiction/uncertain judgments. Following a conservative definition, answers that are either \texttt{contradicted by} or \texttt{not verifiable from} the image are labeled as hallucinations. See Sec.~A.1 for details.
    
\end{itemize}

\subsection*{B.4 Data Preprocessing and Dataset}

We construct a feature dataset with several hyperparameters summarized in Table~\ref{tab:dataset_hyper}.

\begin{table}[h!]
\caption{Hyperparameters for dataset construction.}
\label{tab:dataset_hyper}
\centering
\begin{tabularx}{\columnwidth}{l X}
\toprule
\textbf{Parameter} & \textbf{Value / Description} \\
\midrule
Max token sequence length & 120 \\
Max visual patch sequence length & 512 \\
Max aligned visual sequence length & 512 \\
Feature types & \texttt{ll}, \texttt{ent}, \texttt{emb}, \texttt{mm\_patch}, \texttt{mm\_align} (dimensions depend on the corresponding VLLM architecture) \\
Padding & Zero-padding on the right (sequence dimension) \\
Subsampling & Random selection of 2000 training samples and 1000 test samples per model\\
\bottomrule
\end{tabularx}
\end{table}

\subsection*{B.5 Implementation Details of Branch Encoding and Attention Fusion}
\setcounter{subsubsection}{0}  
\subsubsection{Branch-wise Evidence Encoding}
Each uncertainty source $\mathbf{X}_k \in \mathbb{R}^{L_k \times D_k}$ is encoded independently by a branch module that compresses the sequence into a fixed-dimensional embedding $\mathbf{h}_k \in \mathbb{R}^{d}$ (with $d=64$ in all experiments). The branch encoder supports multiple compression strategies:

\begin{itemize}
    \item \textbf{Linear projection + normalization}: two successive linear layers with intermediate LayerNorm and ReLU activation, followed by global averaging across the sequence.
    \item \textbf{Trainable sequence compressor}: a single linear projection to $d$ dimensions, LayerNorm, ReLU activation, and mean pooling along the sequence.
    \item \textbf{Convolutional encoder}: two 1D convolutional layers (kernel size 3, stride 1), each followed by ReLU activation, and adaptive average pooling to produce a fixed-length embedding.
\end{itemize}

Visual features are aggregated across patches, while token-level features summarize sequence statistics. All branches produce embeddings of the same dimensionality, enabling consistent fusion across heterogeneous sources.

\subsubsection{Cross-branch Attention Fusion}
Given branch embeddings $\mathbf{H} = [\mathbf{h}_1, \dots, \mathbf{h}_K] \in \mathbb{R}^{K \times d}$, cross-branch attention computes a weighted sum:

\begin{itemize}
    \item Each branch embedding is projected into an intermediate attention space of dimension $d_a = 32$.
    \item Non-linear activation ($\tanh$) is applied, followed by a linear scoring to produce unnormalized attention logits.
    \item Attention weights are obtained via softmax across branches and used to compute the weighted sum of embeddings.
\end{itemize}

This mechanism emphasizes informative sources while maintaining contributions from other branches.

\subsubsection{Gated Attention Fusion}
To enhance adaptivity, the aggregated representation $\mathbf{h}$ is further modulated via a sigmoid gate:

\begin{itemize}
    \item A linear projection maps $\mathbf{h}$ to the same embedding dimension $d$.
    \item Sigmoid activation produces element-wise gates.
    \item The final fused embedding is computed as $\mathbf{h} \odot \mathbf{g} + \mathbf{h}$, preserving the residual information.
\end{itemize}

\subsection*{B.6 Training Hyperparameters}

All models are trained using the AdamW optimizer with a weight decay of 0.01.
The learning rate is set to $1 \times 10^{-4}$ with a batch size of 32.
We adopt binary cross-entropy loss as the training objective.
Training is performed for up to 40 epochs with early stopping,
using a patience of 3 epochs based on validation AUROC.
All experiments are trained on two RTX 3090 GPUs in parallel, 
and model inference is performed on a single GPU.

\subsection*{B.7 Evaluation Metrics}

Following previous work~\cite{manakul2023selfcheckgpt,farquhar2024detecting,grewal2024improving,zhao2025sese,tong2025semantic,tong2025halunet}, we report the following metrics (Table~\ref{tab:metrics}) on the \textbf{test} set (or held-out validation set when no official test set exists).

\begin{table}[ht]
\centering
\caption{Main evaluation metrics.}
\label{tab:metrics}
\footnotesize
\renewcommand{\arraystretch}{1.1} 
\begin{tabular}{@{} p{0.28\columnwidth} p{0.65\columnwidth} @{}}
\toprule
Metric & Description \\
\midrule
AUROC & Area under the ROC curve \\
AURAC & Area under the Rejection-Accuracy Curve \\
F1@Best & Highest F1 score on the test set \\
RejAcc@50 & Accuracy after rejecting the 50\% most uncertain samples \\
\bottomrule
\end{tabular}
\end{table}

For our supervised hallucination detection setting, the computation of \textbf{AURAC} and \textbf{RejAcc@50} differs slightly from previous uncertainty-based methods~\cite{manakul2023selfcheckgpt,farquhar2024detecting,grewal2024improving,tong2025semantic}.

Specifically, when using scores generated from supervised methods, the uncertainty of each sample is defined as
\(\mathrm{uncertainty} = 1 - 2 \cdot |p - 0.5|\), 
where \(p \in [0,1]\) is the predicted probability for a positive label (hallucinated).
The most uncertain samples (highest \(\mathrm{uncertainty}\)) are sequentially rejected to construct the \emph{acceptance–rejection curve}.

\textbf{AURAC} is then calculated as the area under this curve, integrating the accuracy of the accepted samples over the acceptance fraction.  
\textbf{RejAcc@50} corresponds to the accuracy computed after rejecting the 50\% of samples with the highest uncertainty.

In contrast, standard uncertainty-based methods typically sort samples directly by predicted score (higher score = higher likelihood of being hallucinated) and construct the acceptance–rejection curve accordingly.  
This modification ensures that both metrics are defined consistently for supervised predictions, allowing direct comparison with standard uncertainty-based methods.

\section*{C. Logistic Regression Baseline}

We implement a classical logistic regression (LR) model as a lightweight baseline for hallucination detection. Implementation details are as follows.

\subsection*{C.1 Input Features}

For each generation instance, scalar token-level signals (log-likelihood $\texttt{ll}$ and entropy $\texttt{ent}$) are summarized via mean and standard deviation. High-dimensional features, including language embeddings ($\texttt{emb}$), visual patches ($\texttt{mm\_patch}$), and cross-modal alignment representations ($\texttt{mm\_align}$), are averaged along the sequence dimension. Principal component analysis (PCA) is applied independently to each high-dimensional feature type, retaining up to 64 components. PCA models are fitted on the training set and reused for validation and test splits. All features are concatenated into a single vector and standardized using z-score normalization. The final output per instance is a 1D feature vector paired with a binary label indicating hallucination.

\subsection*{C.2 Training Procedure}

The LR model is implemented using the \textsc{scikit-learn} library and trained with a class-weighted binary cross-entropy loss to mitigate label imbalance. Key hyperparameters include a maximum of 500 optimization iterations and balanced class weights. The training set is further split into training and validation subsets (ratio 9:1) to fit PCA and evaluate model performance, while feature standardization is applied independently to the training features before fitting the LR model.

\subsection*{C.3 Prediction}

At inference, standardized feature vectors from validation or test instances are passed to the trained LR model to obtain hallucination probability scores, which are then used to compute performance metrics and assess model confidence.

\section*{D. Supplementary Ablation Studies}
\label{sup:ablation_sec}

\begin{table*}[t]
\centering
\caption{
\textbf{Ablation study for Sections D.1 and D.3 across backbones.}
\textbf{FaithSCAN (default, with GA)} denotes the architecture used throughout the main paper.
We first examine the effect of gated attention (GA),
and then investigate the impact of reliability-based sample reweighting (RW).
}
\label{tab:ablation_d1_d2_fullbranches}
\footnotesize
\resizebox{\textwidth}{!}{%
\begin{tabular}{lllcccccccccccc}
\toprule
\multirow{2}{*}{\textbf{Backbone}} & \multirow{2}{*}{\textbf{Method}}
& \multicolumn{4}{c}{\textbf{HalLoc-VQA}}
& \multicolumn{4}{c}{\textbf{HaloQuest}}
& \multicolumn{4}{c}{\textbf{VQA v2}} \\
\cmidrule(lr){3-6} \cmidrule(lr){7-10} \cmidrule(lr){11-14}
& & ROC & RAC & F1@B & RA@50
& ROC & RAC & F1@B & RA@50
& ROC & RAC & F1@B & RA@50 \\
\midrule

\multirow{4}{*}{IB-T5-XL} & FaithSCAN (default, with GA) & \textbf{0.754} & \underline{0.778} & 0.755 & 0.792 & 0.713 & \textbf{0.854} & \underline{0.870} & \textbf{0.885} & \textbf{0.752} & \underline{0.820} & \textbf{0.573} & \textbf{0.858} \\
 & FaithSCAN w/o GA & \textbf{0.754} & \textbf{0.781} & \textbf{0.762} & \underline{0.794} & \underline{0.739} & \underline{0.849} & 0.867 & 0.828 & \underline{0.733} & \textbf{0.828} & 0.558 & \underline{0.832} \\
 & FaithSCAN w/ RW & 0.745 & 0.774 & \underline{0.758} & \textbf{0.798} & \textbf{0.803} & 0.843 & \textbf{0.889} & \underline{0.848} & 0.724 & 0.798 & \underline{0.559} & 0.808 \\

\midrule
\multirow{4}{*}{LLaVA-8B} & FaithSCAN (default, with GA) & \textbf{0.779} & \textbf{0.823} & \textbf{0.680} & \textbf{0.873} & 0.827 & 0.848 & \underline{0.762} & \textbf{0.902} & \textbf{0.685} & \underline{0.801} & \textbf{0.494} & \textbf{0.831} \\
 & FaithSCAN w/o GA & \underline{0.723} & 0.741 & \underline{0.570} & 0.804 & \textbf{0.842} & \underline{0.851} & \textbf{0.768} & \textbf{0.902} & \underline{0.660} & 0.763 & \underline{0.462} & 0.780 \\
 & FaithSCAN w/ RW & 0.710 & \underline{0.751} & 0.561 & \underline{0.810} & \underline{0.835} & \textbf{0.853} & 0.758 & 0.893 & 0.655 & \textbf{0.814} & 0.452 & \underline{0.810} \\

\midrule
\multirow{4}{*}{Qwen3-VL-8B} & FaithSCAN (default, with GA) & \underline{0.739} & 0.757 & \underline{0.569} & 0.811 & \underline{0.829} & \underline{0.941} & \textbf{0.532} & \underline{0.975} & \textbf{0.695} & \textbf{0.876} & 0.420 & \textbf{0.884} \\
 & FaithSCAN w/o GA & 0.721 & \underline{0.777} & 0.555 & \underline{0.834} & \textbf{0.832} & \textbf{0.943} & \underline{0.512} & \textbf{0.980} & \underline{0.681} & \underline{0.817} & \underline{0.423} & \underline{0.816} \\
 & FaithSCAN w/ RW & \textbf{0.754} & \textbf{0.818} & \textbf{0.578} & \textbf{0.848} & 0.827 & 0.939 & 0.505 & \underline{0.975} & 0.676 & 0.762 & \textbf{0.433} & 0.754 \\

\bottomrule
\end{tabular}
}
\end{table*}

\subsection*{D.1 Implementation of the Attention Mechanism}
\label{subsec:attention_impl}

We study two variants of cross-branch attention, with results reported in Table~\ref{tab:ablation_d1_d2_fullbranches}. 
In most settings, gated attention yields slightly better performance than the non-gated variant, suggesting that gating enables selective amplification or attenuation of cross-branch evidence and improves representation flexibility.

\setcounter{subsubsection}{0}  
\subsubsection{Attention without Gating} 
Each evidence branch produces an embedding $\mathbf{h}_k \in \mathbb{R}^{d}$, and the set of embeddings is $\mathbf{H} = [\mathbf{h}_1, \dots, \mathbf{h}_K] \in \mathbb{R}^{K \times d}$. Attention weights are computed as
\begin{equation}
\alpha_k = 
\frac{\exp\!\left(\mathbf{w}_a^\top \tanh(\mathbf{W}_a \mathbf{h}_k)\right)}
{\sum_{j=1}^{K} \exp\!\left(\mathbf{w}_a^\top \tanh(\mathbf{W}_a \mathbf{h}_j)\right)},
\qquad
\mathbf{h} = \sum_{k=1}^{K} \alpha_k \mathbf{h}_k,
\end{equation}
where $\mathbf{w}_a$ and $\mathbf{W}_a$ are learnable parameters. This variant does not include a gated residual, serving as a baseline for evaluating the effect of feature-wise modulation.

\begin{figure*}[t]
\centering
\includegraphics[width=\textwidth]{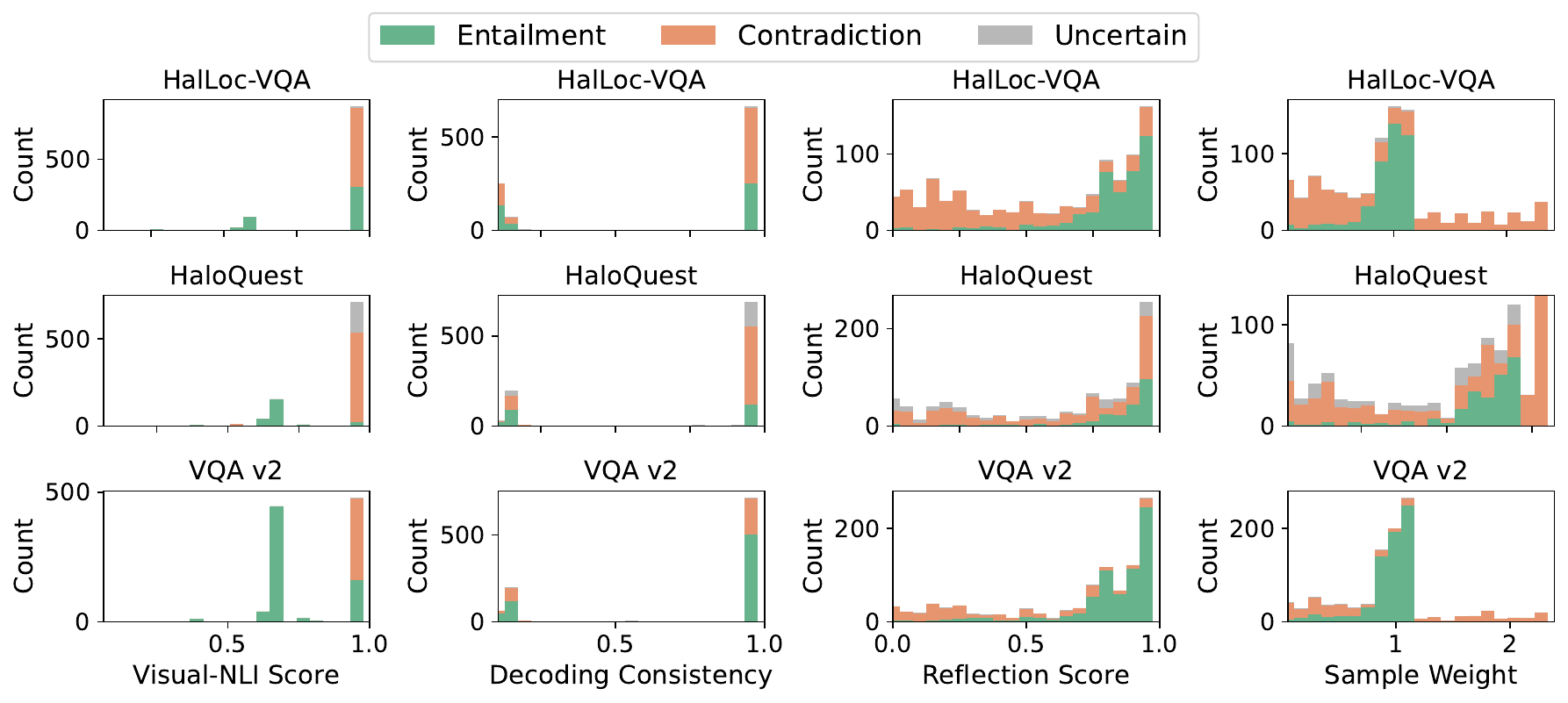}
\caption{Distributions of post-hoc reliability scores and reweighted sample weights on three datasets (IB-T5-XL).}

\label{fig:automated_eval_instruct}
\end{figure*}

\subsubsection{Attention with Gating} 
The gated attention variant is as described in the main manuscript (Sec.~IV-E). For completeness, it applies a gated residual to the attention-fused representation $\mathbf{h}$, producing $\widetilde{\mathbf{h}}$, which allows adaptive feature-wise modulation.

\subsubsection{Key Difference} 
The only difference between the two variants is the gated residual transformation: the ungated version relies solely on attention weights for aggregation, while the gated version additionally modulates each feature of the fused representation through a learnable gate.

\subsection*{D.2 Post-hoc Verification for Reliability}
\label{subsec:posthoc_verification}

\begin{figure*}[t]
\centering
\includegraphics[width=\textwidth]{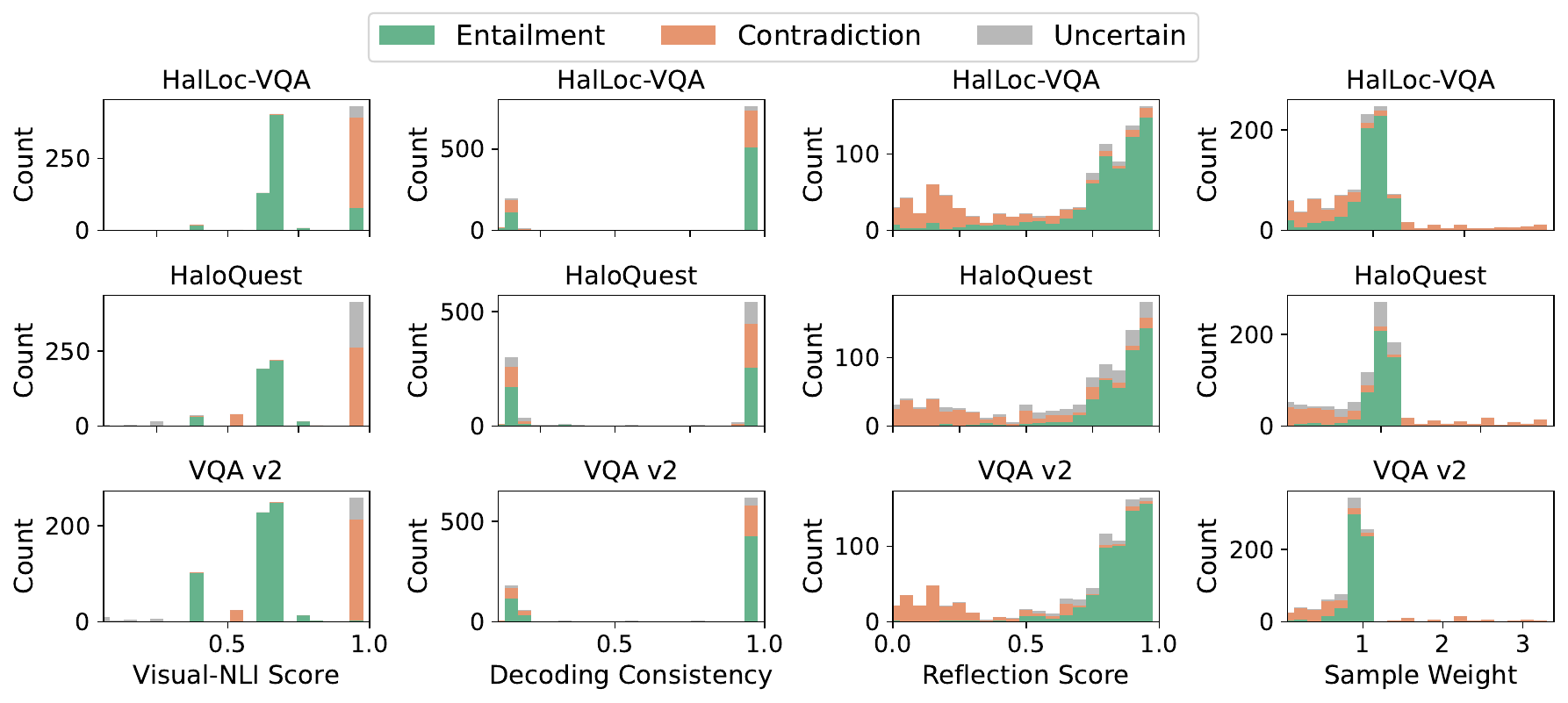}
\caption{Distributions of post-hoc reliability scores and reweighted sample weights on three datasets (LLaVA-8B).}

\label{fig:automated_eval_llava}
\end{figure*}

Given an image--question--answer triplet $(I, q, \hat{a})$ with a model-driven hallucination label $y_\text{hall}$, we compute three complementary post-hoc verification signals using the same vision--language model. All verification is performed \emph{after} answer generation under the same image--question context.

\begin{itemize}
    \item \textbf{Visual-NLI decisiveness.}  
    Let $p_\text{ent}, p_\text{con}, p_\text{unc}$ denote entailment, contradiction, and uncertainty probabilities predicted by Visual-NLI on the generated answer. Averaging over multiple NLI rounds gives $\bar{\mathbf{p}} = (\bar p_\text{ent}, \bar p_\text{con}, \bar p_\text{unc})$. The decisiveness score is:
    \begin{equation}
        s_\text{nli} = 1 - \frac{H(\bar{\mathbf{p}})}{\log 3} \in [0,1],
    \end{equation}
    where $H(\cdot)$ is the Shannon entropy. Higher $s_\text{nli}$ indicates more confident NLI judgments.

    \item \textbf{Stochastic decoding consistency.}  
    Alternative answers are sampled under high-temperature decoding. For each sample $m$, define a soft hallucination score $h^{(m)} = p_\text{con}^{(m)} + p_\text{unc}^{(m)}$. Consistency is quantified as:
    \begin{equation}
        s_\text{stoch} = \exp\Big(- \operatorname{Var}\big(\{h^{(m)}\}_{m=1}^{M}\big)\Big),
    \end{equation}
    which decreases when stochastic outputs disagree.

    \item \textbf{Reflection-based self-consistency.}  
    The model re-evaluates its own answer with visual evidence, producing reflection scores $\{r^{(t)}\}_{t=1}^{T}$. We define:
    \begin{equation}
        s_\text{ref} =
        \begin{cases}
            \displaystyle \max(\text{min\_s\_ref}, 1 - \frac{1}{T} \sum_{t=1}^T r^{(t)}), & y_\text{hall} = 1, \\
            \displaystyle \max(\text{min\_s\_ref}, \frac{1}{T} \sum_{t=1}^T r^{(t)}), & y_\text{hall} = 0,
        \end{cases}
    \end{equation}
    with $\text{min\_s\_ref}=0.05$ to prevent zero values.
\end{itemize}

These scores $\{s_\text{nli}, s_\text{stoch}, s_\text{ref}\}$ provide independent measures of the reliability of the hallucination label.

\begin{figure*}[t]
\centering
\includegraphics[width=\textwidth]{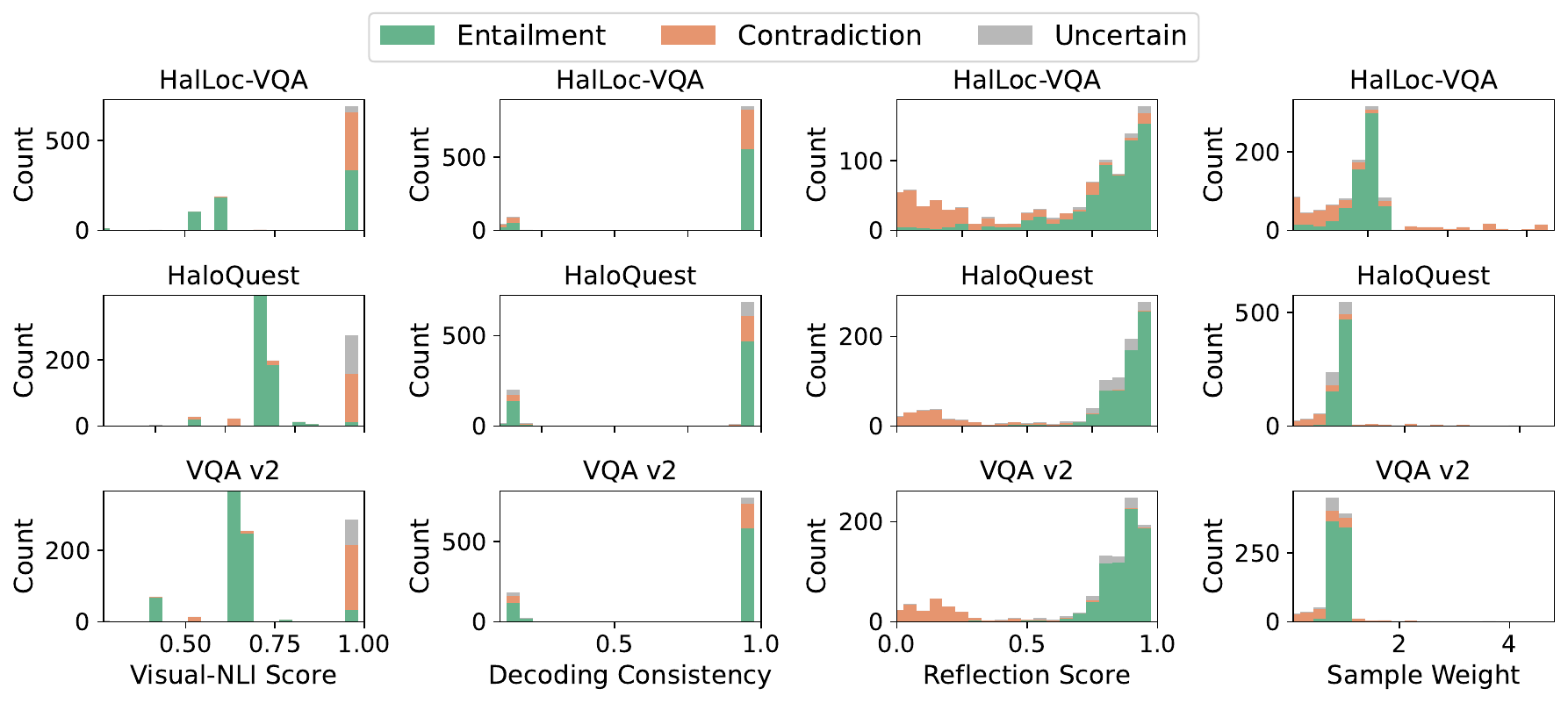}
\caption{Distributions of post-hoc reliability scores and reweighted sample weights on three datasets (Qwen3-VL-8B).}

\label{fig:automated_eval_qwen}
\end{figure*}

We report the distributions of reliability scores produced by the three post-hoc verification methods across different vision--language model backbones, with results shown in Figs~\ref{fig:automated_eval_instruct}, \ref{fig:automated_eval_llava}, and \ref{fig:automated_eval_qwen}. 
Overall, the observed trends are consistent with those reported in the main paper, indicating that the proposed verification signals provide stable and meaningful reliability estimates across models.

Notably, stronger VLM backbones tend to exhibit lower hallucination rates, which is reflected by a higher concentration of samples with confident and reliable verification scores. 
This observation suggests that model capacity and instruction-following ability play an important role in reducing hallucinations, while our post-hoc verification framework remains effective across a wide range of model strengths.

\subsection*{D.3 Reliability-Aware Supervision}
\label{subsec:weighted_supervision}

To incorporate post-hoc verification into training, we combine the raw reliability signals into a single scalar weight per instance:
\begin{equation}
    w_i^\text{raw} = \lambda_\text{nli} s_\text{nli} +
                      \lambda_\text{stoch} s_\text{stoch} +
                      \lambda_\text{ref} s_\text{ref} \in [0,1],
\end{equation}
where $\lambda_\text{nli}$, $\lambda_\text{stoch}$, and $\lambda_\text{ref}$ control the contribution of each verification signal. 
In our experiments, we set $\lambda_\text{nli}=0$, $\lambda_\text{stoch}=0$, and $\lambda_\text{ref}=1$, effectively using the reflection-based signal as the sole reliability score.

To ensure that the weights are comparable across classes, we normalize them within each class so that the average weight for each class equals 1. 
Let $w_i^\text{raw}$ denote the initial weight of sample $i$, and $y_i\in\{0,1\}$ its label (1 for hallucination, 0 for non-hallucination). 
The normalized weight $w_i$ is then computed as
\begin{equation}
w_i =
\begin{cases}
\dfrac{w_i^\text{raw}}{\frac{1}{N_\text{pos}} \sum_{j:y_j=1} w_j^\text{raw}}, & y_i=1,\\[2mm]
\dfrac{w_i^\text{raw}}{\frac{1}{N_\text{neg}} \sum_{j:y_j=0} w_j^\text{raw}}, & y_i=0,
\end{cases}
\end{equation}
where $N_\text{pos}$ and $N_\text{neg}$ are the number of samples in the positive (hallucination) and negative (non-hallucination) classes, respectively. 
After this normalization, the average weight within each class is exactly 1.

FaithSCAN predicts the hallucination probability $p_i$ using a reliability-weighted binary cross-entropy loss:
\begin{equation}
    \mathcal{L} = \frac{1}{N} \sum_{i=1}^{N} w_i \Big[-y_i \log p_i - (1-y_i) \log (1-p_i) \Big].
\end{equation}

This design allows highly reliable samples to contribute more during training while down-weighting ambiguous or weakly supported instances. It mitigates the impact of noisy labels and does not introduce additional parameters or inference-time overhead.

The last column of Figs.~\ref{fig:automated_eval_instruct}, \ref{fig:automated_eval_llava}, and \ref{fig:automated_eval_qwen} illustrates the distributions of sample weights after reweighting, and the corresponding experimental results are reported in Table~\ref{tab:ablation_d1_d2_fullbranches}. 

The results show that reliability-aware reweighting affects samples selectively. The weights of non-hallucinated samples remain largely unchanged, suggesting that these samples are generally stable and do not require adjustment. In contrast, a small portion of hallucinated samples with high reliability receive substantially increased weights.

Training with these reweighted samples leads to improvements under certain settings, highlighting the potential of this approach as a promising direction for future research.

\section*{E. Explainability via Gradient $\times$ Input}
To better understand the contributions of different feature channels (e.g., token-level log-likelihood \texttt{ll}, entropy \texttt{ent}, and embedding features \texttt{emb}) in hallucination detection, we adopt a gradient-based explainability approach, inspired by classic saliency analysis~\cite{simonyan2014deep,shrikumar2017learning}.

Let the input feature tensor be $\mathbf{x} = \{x_{t,f}\}$, where $t$ indexes tokens and $f$ indexes feature types. Let the model be $f_\theta(\mathbf{x})$, and the hallucination probability output be $\hat{y} = \sigma(f_\theta(\mathbf{x}))$. The attribution for each token-feature pair is computed as:

\[
A_{t,f} = \left(\frac{\partial \hat{y}}{\partial x_{t,f}}\right) \odot x_{t,f},
\]

where $\frac{\partial \hat{y}}{\partial x_{t,f}}$ is the gradient of the predicted hallucination probability with respect to the input feature $x_{t,f}$, and $\odot$ denotes element-wise multiplication.

Intuitively, the gradient $\frac{\partial \hat{y}}{\partial x_{t,f}}$ captures the local sensitivity of the model output to changes in the feature, while the feature value $x_{t,f}$ represents the actual signal strength at the input. Their element-wise product captures both the magnitude and direction of the feature's contribution to the model's decision.

To obtain a token-level contribution score, we aggregate along the feature dimension:

\[
C_{t,f} = \sum_{d} A_{t,f,d}.
\]

If $C_{t,f} > 0$, it indicates that this token's feature has a \textbf{positive effect} on hallucination prediction (i.e., increasing this feature increases the probability of predicting hallucination). Conversely, if $C_{t,f} < 0$, it indicates a \textbf{negative effect}.

For visualization, we plot the $C_{t,f}$ sequence for each token and feature type $f \in \{\texttt{ll}, \texttt{ent}, \texttt{emb}\}$, providing a token-level decomposition of the model's decision. This analysis reveals which linguistic or uncertainty features drive the hallucination detection predictions.
\end{document}